\documentclass[lettersize,journal]{IEEEtran}
\usepackage{amsmath,amsfonts}
\usepackage{algorithmic}
\usepackage{algorithm}
\usepackage{array}
\usepackage[caption=false,font=normalsize,labelfont=sf,textfont=sf]{subfig}
\usepackage{textcomp}
\usepackage{stfloats}
\usepackage{url}
\usepackage{verbatim}
\usepackage{graphicx}
\usepackage{cite}
\usepackage{xcolor}
\usepackage{hyperref}
\usepackage[numbers,sort&compress]{natbib} 

\usepackage{tikz}
\usetikzlibrary{calc}

\usepackage{lipsum}  
\usepackage{float}

\begin{document}

\title{Infrastructure-based Autonomous Mobile Robots for Internal Logistics - Challenges and Future Perspectives}


\author{Erik Brorsson$^{1,2}$, Kristian Ceder$^1$, Ze Zhang$^1$, Sabino Francesco Roselli$^1$, Endre Erős$^3$, Martin Dahl$^1$, Beatrice Alenljung$^4$, Jessica Lindblom$^5$, Thanh Bui$^6$, Emmanuel Dean$^1$, Lennart Svensson$^1$, Kristofer Bengtsson$^2$, Per-Lage Götvall$^2$, Knut Åkesson$^1$
\thanks{\noindent $^1$Dept. of Electrical Engineering, Chalmers University of Technology, Gothenburg, Sweden.\\  $^2$Global Trucks Operations, Volvo Group, Gothenburg, Sweden. \\ $^3$Chalmers Industriteknik, Gothenburg, Sweden. \\ $^4$School of Informatics, University of Skövde, Skövde, Sweden. \\ $^5$Dept. of Information Technology, Uppsala University, Uppsala, Sweden. \\$^6$RISE Research Institutes of Sweden, Gothenburg, Sweden. \\Corresponding author: erikbro@chalmers.se.}
}



\maketitle

\begin{tikzpicture}[remember picture, overlay]
\node[anchor=north west, xshift=10pt, yshift=-10pt] at (current page.north west) {
    \parbox{\dimexpr\textwidth-20pt\relax}{
        \footnotesize
        This work has been submitted to the IEEE for possible publication. 
        Copyright may be transferred without notice, after which this version 
        may no longer be accessible.
    }
};
\end{tikzpicture}


\begin{figure}
    \centering
    \includegraphics[width=0.8\linewidth]{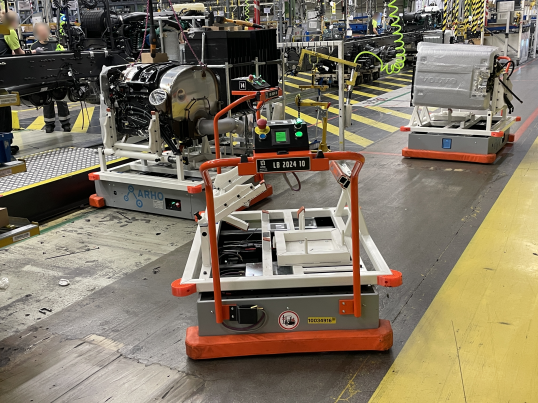}
    \caption{Three of the autonomous transport robots operating inside the factory.}
    \label{fig:robot}
\end{figure}

    

The adoption of Autonomous Mobile Robots (AMRs) for internal logistics is accelerating, with most solutions emphasizing decentralized, onboard intelligence. While AMRs in indoor environments like factories can be supported by infrastructure, involving external sensors and computational resources, such systems remain underexplored in the literature.
This paper presents a comprehensive overview of infrastructure-based AMR systems, outlining key opportunities and challenges. To support this, we introduce a reference architecture combining infrastructure-based sensing, on-premise cloud computing, and onboard autonomy. Based on the architecture, we review core technologies for localization, perception, and planning. We demonstrate the approach in a real-world deployment in a heavy-vehicle manufacturing environment and summarize findings from a user experience (UX) evaluation. Our aim is to provide a holistic foundation for future development of scalable, robust, and human-compatible AMR systems in complex industrial environments.

\section{Introduction}
The demand for automation in internal logistics is steadily increasing across a wide range of industrial sectors. While significant progress has been made in structured settings, such as fully automated warehouses, achieving effective automation in less regulated and dynamic environments remains a considerable challenge. 
For example, at Volvo Trucks' final assembly plants, the production process must accommodate a broad range of vehicle types, including battery-electric, internal combustion engine, and hydrogen-powered trucks. These variants are all assembled on the same production line, which significantly increases system complexity and requires a high degree of operational flexibility. This variability complicates automation efforts and reinforces the industry's reliance on human workers. Nevertheless, as automation technologies continue to mature, they are progressively introduced to complement the human workforce. In these environments, we can expect that humans and robots will work side by side, or in a collaborative sense, for the foreseeable future.

The need for flexibility in material handling has driven the development of Autonomous Mobile Robots (AMRs) \cite{fragapane2021planning}. AMRs are industrial, typically wheeled, robots that transport materials in dynamic, human-shared environments through decentralized decision-making and onboard navigation. Current solutions rely on onboard sensors, such as LiDAR and cameras, to perform simultaneous localization and mapping (SLAM) and detect obstacles in real-time for collision-free motion. Although these systems have seen substantial progress in recent years \cite{fragapane2021planning, bernardo2022survey, lackner2024review, fottner2021autonomous, niloy2021critical, rubio2019review, alatise2020review}, several challenges remain. Generally, onboard sensors' limited range and susceptibility to occlusion hinder reliable perception in crowded environments and far away from the robot. SLAM is further complicated by moving objects that interfere with the mapping process. Additionally, mobile platforms are constrained by limited computational resources, which prohibits the use of computationally intensive algorithms for perception and decision making. These issues severely limit the robustness and reliability of AMR systems in complex indoors environments.

To overcome these limitations, recent research has proposed AMR systems that leverage cloud computing and infrastructure-based sensing under paradigms such as Cloud Robotics \cite{hu2012cloud} and Internet of Robotic Things (IoRT) \cite{simoens2018internet}. This offers multiple benefits, such as enhanced perception capabilities thanks to additional sensors and joint data processing, increased computational budget at the cloud or edge-clusters, and global coordination across a fleet of robots. 
Notable examples include: offloading time-critical algorithms for perception and planning to the cloud \cite{balogh2021cloud}, collaborative SLAM with multiple robots \cite{arumugam2010davinci, riazuelo2014c2tam, karrer2018cvi, mohanarajah2015cloud}, and the use of infrastructure-mounted cameras for localization \cite{alam2024fiducial}, perception \cite{zhang2018autonomous, alam2024fiducial}, or navigation \cite{lee2003controlling, reina2014zeppelin, streit2016vision}. 

However, the integration of infrastructure and cloud resources into AMR systems is still an emerging area of research. Recent surveys continue to focus primarily on onboard autonomy, and existing work involving infrastructure tends to address isolated functions or application-specific setups. A cohesive systems-level perspective is currently lacking. To help address this gap, we first propose a reference architecture for AMR systems that incorporates infrastructure-mounted sensors, cloud computing, and onboard intelligence to support a wide range of applications. Moreover, while key AMR capabilities such as localization, perception, and planning have been extensively studied in onboard-only systems, their design and performance are fundamentally different in infrastructure-supported setups. We therefore review these core functions from the perspective of the proposed architecture to identify emerging research opportunities and design considerations.

To ground our discussion, we also present a real-world industrial deployment of an infrastructure-based AMR system and summarize insights from its evaluation. While prior work often remains conceptual or limited to lab settings, our implementation in a live factory environment highlights practical benefits and challenges. Finally, drawing from both literature and industrial experience, we identify critical open challenges and suggest promising directions for future research.

Our contributions can be summarized as follows:
\begin{itemize}
    \item We propose a reference architecture for AMR systems that integrates infrastructure sensors and on-premise cloud computing resources.
    \item We review enabling technologies for key AMR functions, including localization, perception, and planning.
    \item We present an industrial deployment of the system and summarize key user experience (UX) evaluation results.
    \item Based on our review and evaluation, we identify current challenges and outline promising research directions.
\end{itemize}

The remainder of this paper is organized as follows. Section II defines the problem and outlines the reference architecture. Section III reviews essential supporting technologies. Section IV presents our industrial evaluation. Section V discusses open research challenges and future directions. Finally, Section VI concludes the paper.

\section{Automated Transports for Internal Logistics}

\subsection{Problem formulation}
\label{sec:problem}
We address the problem of autonomously managing transportation tasks in a factory using a fleet of wheeled autonomous mobile robots (AMRs). The task set is defined as $\mathcal{T}=\{\tau_i \}_i$, where a task $\tau_i=[A,B, T_A, T_B]$ consists of a start location $A$, an end location $B$, and associated time windows $T_A$ and $T_B$. Specifically, $T_A= [t_{A_0}, t_{A_1}]$ defines the permissible arrival window at location $A$, while $T_B = [t_{B_0}, t_{B_1}]$ defines the corresponding window at location $B$.
The factory environment is dynamic due to uncontrollable events such as pedestrian movement and manually operated forklifts. At the same time, it is regulated by factory-specific traffic rules and conventions. Some are explicitly marked, for example with floor signage, while others are implicit through established practice. These rules define where robots may drive and which vehicle has priority at intersections. The objective is to control the AMR fleet to safely and efficiently complete the transportation tasks while complying with these rules.

Achieving this objective requires a system capable of perceiving and understanding its environment. First, the system must estimate the current state, including the locations of static and dynamic obstacles and the positions of all AMRs. As sensor data is noisy and incomplete, this estimate is inherently uncertain. To enable proactive behavior, the system must also predict likely future events, such as pedestrian or vehicle movements. These predictions are likewise uncertain and may change rapidly with human behavior. Developing an adequate understanding of the current and future state of the environment is therefore a key perception challenge that requires robust processing and interpretation of sensor data.

Beyond perception, the system must plan the actions of all AMRs. At the highest level, task assignment, scheduling, and route selection grows rapidly in complexity with the number of tasks, robots and routing alternatives, which makes it difficult to solve for larger instances. During execution, each robot must continuously select safe and effective actions to progress toward its goals while navigating near pedestrians and other vehicles. In crowded indoor settings, this involves reasoning about traffic rules and human intentions and may also require cooperation to avoid deadlocks. For example, some vehicles, such as factory tugger trains, cannot reverse, which may require strategic maneuvering from nearby AMRs. In summary, the system must respond quickly to environmental changes while maintaining predictable behavior that supports safe and effective human–robot collaboration, which pose a significant planning challenge.

\subsection{Reference Architecture}
        \begin{figure*}[h!]
            \centering
            \includegraphics[width=0.85\linewidth]{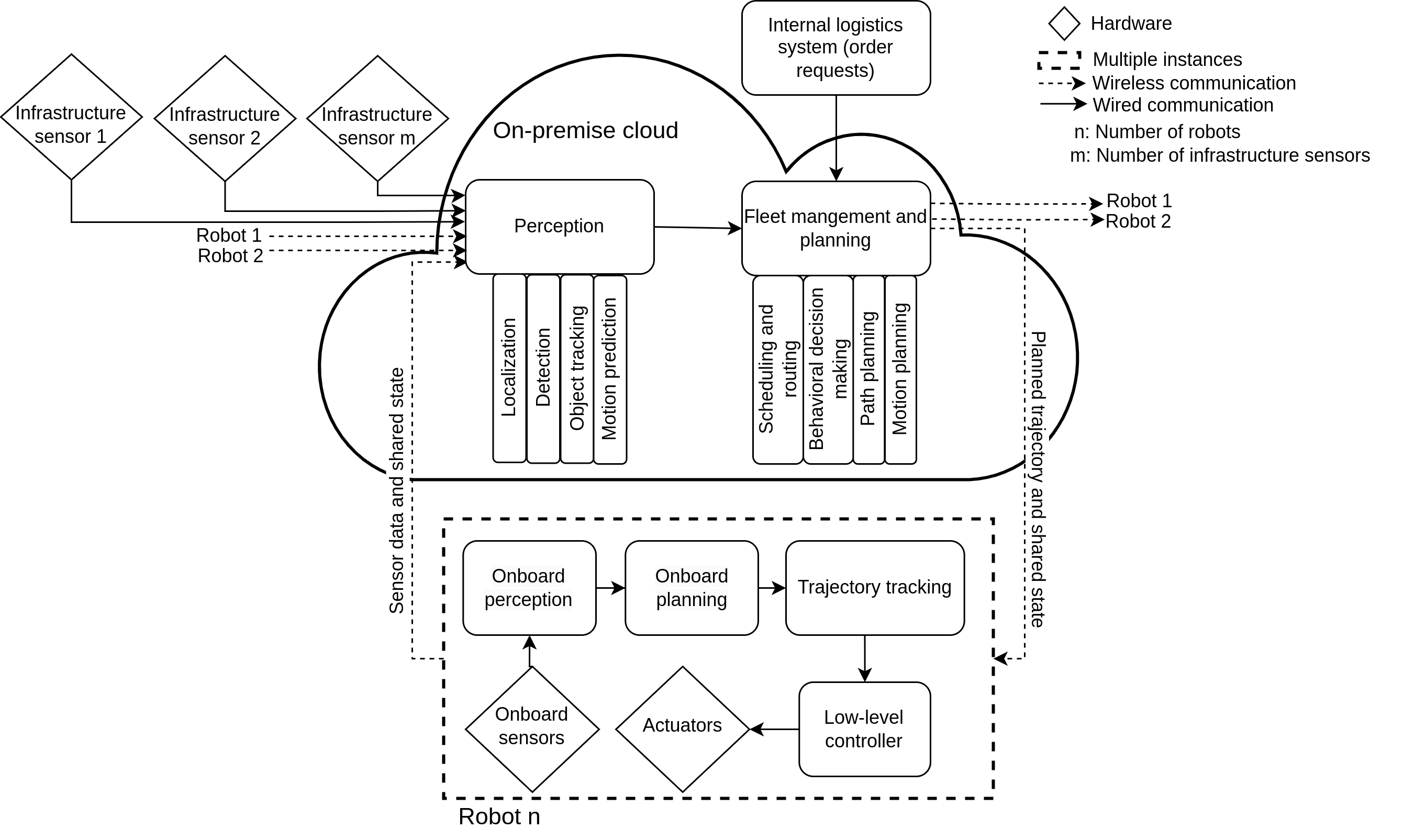}
            \caption{RAIL:\textit{ a \textbf{R}eference \textbf{A}rchitecture for \textbf{I}nfrastructure-based AMR Systems in Internal \textbf{L}ogistics}.}
            \label{fig:ref_arch}
        \end{figure*}
A common approach to autonomous navigation in dynamic environments is a modular software architecture that separates perception and decision-making components \cite{rubio2019review, alatise2020review, zhang2023automated, lackner2024review, badue2021self, yurtsever2020survey, zhao2024autonomous}.
Perception is typically divided into localization, which estimates the robot’s position, and environment perception. The latter includes estimating the current environment state (e.g., road layout and obstacle positions), tracking dynamic objects such as other road users, and predicting their future trajectories \cite{zhao2024autonomous}. In our architecture, which is illustrated in Figure \ref{fig:ref_arch}, we summarize the perception module by localization, detection, object tracking, and motion prediction.

Based on the perception results, the decision-making module provides a long-term route to the destination, makes short-to-mid-term decisions to progress toward it, and compute a feasible trajectory to execute the chosen behavior. In self-driving cars, this pipeline may be described by route planning, path planning, behavior selection, and motion planning \cite{badue2021self}. By adding a task scheduler that allocates transportation tasks across the AMR fleet, the same structure applies to our application.
Accordingly, we describe the decision making system as scheduling and routing, behavioral decision making, path planning, and motion planning in Figure \ref{fig:ref_arch}. The functionality of each perception and decision-making component is detailed in the next section.

Traditionally, both perception and decision-making have been performed entirely onboard, relying on sensors such as cameras, LiDAR, and radar. However, a growing trend in both autonomous vehicles and robotics is to leverage infrastructure and increase connectivity between agents. In autonomous driving, for example, vehicle-to-everything (V2X) communication is increasingly used for shared perception and cooperative planning \cite{hakak2023autonomous}.
Similarly, the robotics community has embraced cloud-based approaches in which key capabilities are offloaded from the robot to external infrastructure.
For example, \cite{balogh2021cloud} demonstrates a fully cloud-offloaded navigation stack, where onboard sensor data is transmitted to a 5G-enabled edge cloud for real-time perception and control, with resulting commands sent back to the AMR. Other systems \cite{arumugam2010davinci, riazuelo2014c2tam, karrer2018cvi, mohanarajah2015cloud} focus on cloud-based SLAM execution with multiple robots. Infrastructure-mounted cameras have also been used for localization and guidance \cite{lee2003controlling, reina2014zeppelin, streit2016vision}, with perception and planning executed directly on the camera hardware. Ceiling-mounted cameras further serve as auxiliary sensors to improve localization \cite{alam2024fiducial} and perception \cite{zhang2018autonomous, alam2024fiducial}.

To leverage the benefits of these methods, we propose RAIL:\textit{ a \textbf{R}eference \textbf{A}rchitecture for \textbf{I}nfrastructure-based AMR Systems in Internal \textbf{L}ogistics}. As illustrated in Figure \ref{fig:ref_arch}, RAIL generalizes and extends existing AMR systems by enabling a broad range of cloud-based capabilities, such as collaborative perception between robots, enhanced sensing through infrastructure-mounted devices, and access to centralized, high-performance computing. 
In RAIL, AMRs and external sensors (e.g., ceiling-mounted cameras) transmit sensor data or extracted features to the on-premise cloud, which serves as a central hub for global perception and fleet coordination. Based on transportation requests from the internal logistics system, the cloud platform computes task schedules and motion plans in the form of trajectories, which are communicated wirelessly to the fleet. Additional information, such as environment maps, robot states, and progress of ongoing transportation tasks, may also be shared between the AMRs and the on-premise cloud. In Figure \ref{fig:ref_arch}, we simply denote such information as the \textit{shared state}.
Each AMR uses a local trajectory tracker and low-level controller to execute the received motion plans.

While the architecture is designed to capitalize on cloud-based intelligence, it also accommodates onboard autonomy where needed. In some environments, onboard sensing may be the only viable source of accurate, close-range information. Furthermore, onboard capabilities ensure robustness to intermittent connectivity or infrastructure failure. Depending on the specific application and reliability requirements, the onboard stack may range from minimal functionality (e.g., line following) to fully developed autonomous navigation.

\section{Key Technologies}

    In this section, we review the key components of the reference architecture in Figure \ref{fig:ref_arch}. Specifically, the following subsections are dedicated to localization, environment perception, and fleet management and planning. 

    \subsection{Localization}

For AMRs, localization involves estimating the robot’s current position and orientation within its environment. Accurate localization is a fundamental prerequisite for enabling robots to navigate, perform tasks, and interact safely with their surroundings. While onboard sensors such as inertial measurement units (IMUs) and odometry can track the robot’s motion in a local coordinate frame over short durations, additional sensing is required to determine the robot’s global position. Unlike outdoor settings, where Global Navigation Satellite Systems (GNSS) are widely used, reliable GNSS signals are typically unavailable indoors. Consequently, various alternative indoor localization methods have been developed.

The most common approach for indoor localization is Simultaneous Localization and Mapping (SLAM) \cite{huang2023indoor}. SLAM enables a robot to operate in unknown environments by using onboard sensors to simultaneously construct a map and localize itself within it. This problem has been extensively studied in the literature \cite{cadena2017past, zou2021comparative, macario2022comprehensive} and is supported by widely used open-source tools, such as the ROS 2 Navigation Stack \cite{macenski2023survey, macenski2021slam}. 
SLAM techniques have been developed for both LiDAR sensors (e.g., GMapping \cite{grisetti2007improved}, KartoSLAM \cite{konolige2010efficient}, and Cartographer \cite{hess2016real}) and camera-based systems, including monocular, stereo, and RGB-D setups \cite{campos2021orb}. Given the high computational demands of SLAM, cloud-based solutions have been proposed to offload processing from the robot itself \cite{balogh2021cloud}. Such approaches not only benefit individual robots but also support collaborative mapping in multi-robot systems \cite{karrer2018cvi, riazuelo2014c2tam, mohanarajah2015cloud}. Although recent advancements in SLAM has led to increased robustness and accuracy, SLAM-based localization remains challenging in highly dynamic environments.

In structured indoor environments, dedicated localization infrastructure is often installed to simplify the problem. Radio-frequency-based technologies such as RFID, Ultra-Wideband (UWB), Wi-Fi, Bluetooth, and Zigbee provide low-cost positioning, though typically with limited accuracy \cite{huang2023indoor}. Combining such systems with SLAM or odometry/IMU data is a promising approach to enhance robustness and reliability \cite{fragapane2021planning, panigrahi2022localization}. Additional methods involve placing artificial landmarks, such as LEDs, reflectors, or fiducial markers, within the environment, which the robot can detect using its onboard sensors \cite{bernardo2022survey}. Localization is then achieved by estimating the bearing or distance to these targets. Another alternative is to use infrastructure cameras to monitor the robot from a fixed viewpoint. In such systems, a fiducial marker is often attached to the robot to facilitate accurate pose estimation \cite{song2018design, alam2024fiducial}.

    \subsection{Environment Perception}
        

    \begin{figure}[h!]
        \centering
        \includegraphics[width=1\linewidth]{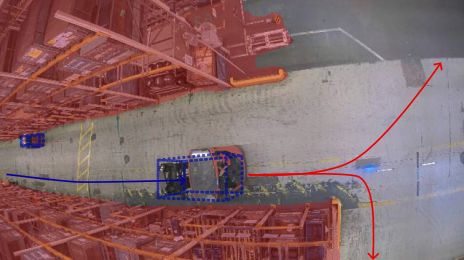}
        \caption{Example perception results in a factory environment. Semantic segmentation is used to find static obstacles (red) and bounding box detection is used for vehicles (blue). Object tracking estimates the past trajectory of the vehicle (blue arrow), and motion prediction creates multiple hypothesis for the future movement (red arrows).} 
        \label{fig:perception}
    \end{figure}

    Perception involves estimating the state the environment, such as the 3D positions and velocities of nearby vehicles and the road layout. Depending on the task, both parametric representations (e.g., bounding boxes and splines) or non-parametric representations (e.g., occupancy maps and voxel grids) are used. 
    In the context of internal logistics, where both the robot and surrounding agents are restricted to planar motion, a bird’s-eye view (BEV) representation is particularly well suited. 
    Beyond estimating the current state of the environment, perception also involves reasoning about possible future states. This foresight enables the robot to make proactive decisions that enhance both safety and operational efficiency. Figure \ref{fig:perception} illustrates example output of the perception module consisting of static obstacles estimation alongside detection, tracking and motion prediction of dynamic objects.  
    In the remainder of this section, we review commonly used perception methods described in the literature. While not universal, many of these approaches follow a modular pipeline that includes detection, tracking, and motion prediction.

    \subsubsection{Detection}

 



        Most literature on perception for AMRs and autonomous vehicles focus on onboard systems based on e.g., cameras, LiDAR and radar sensors \cite{zhao2024autonomous, liu2020computing, yurtsever2020survey}. 
        Deep learning methods currently dominate most computer vision tasks, including object detection, depth estimation, and all types of scene segmentation (semantic, instance, panoptic), regardless of the specific sensor configuration. 
        While progress is made on each of the mentioned tasks, creating a rich and dense representation of the environment remains challenging. Here, BEV perception has emerged as a promising solution, where e.g., multiple cameras \cite{philion2020lift}, LiDAR \cite{lang2019pointpillars} or a combination of both \cite{li2024bevformer, liang2022bevfusion}, are used to aggregate features on a virtual ground plane. Such a feature representation enables subsequent BEV segmentation and object detection, which can easily be consumed by downstream planning modules \cite{li2023delving}.
        Despite the significant progress made, onboard perception remains vulnerable to occlusions and obstructions, leading to degraded performance in crowded environments.

        Infrastructure-based perception literature mainly revolve around surveillance applications, where cameras are used to monitor environments such as traffic intersection or indoor facilities.
        Since the goal is to detect objects on the ground/floor, which is typically relatively flat, these applications benefit from simplified image-to-world coordinate mapping via precomputed calibration. Early methods used probabilistic frameworks to directly infer the ground plane positions based on detections in the images and precomputed transformation \cite{fleuret2008multicamera, roig2011conditional}. Recent works, however, learn dense BEV representations of the environment, which is particularly well suited for fusing information from multiple cameras viewing the same area \cite{hou2020multiview}. 


        With the advent of reliable, low-latency communication technologies, such as 5G, vehicle-to-vehicle (V2V) and vehicle-to-everything (V2X) perception has gained traction. These paradigms rely on information sharing between multiple autonomous agents and/or infrastructure devices, such as roadside cameras or LiDAR units, to overcome key limitations of onboard sensing, such as restricted range and blind spots \cite{liu2023towards}. 
        In the robotics domain, multiple robots have been used for collaborative mapping of the environment \cite{karrer2018cvi, riazuelo2014c2tam, mohanarajah2015cloud}.
        Recent V2V applications in autonomous driving involve using multi-view images or LiDAR from connected vehicles to perform 3D object detections \cite{hu2023collaboration, pmlr-v205-li23e, li2022learningdistilledcollaborationgraph}, BEV segmentation \cite{pmlr-v205-li23e}, or to predict the semantic occupancy status of 3D voxels \cite{song2024collaborative}.
        In terms of V2X communication, \cite{xu2022v2x} proposes a method for 3D object detection based on LiDAR observations from two connected vehicles and one roadside unit. 
        While promising, these technologies are still in early development. Most current systems are limited to small-scale setups, and extensions to multiple vehicles and roadside units are rare. Moreover, creating training data for these systems is particularly challenging and expensive, as it typically requires detailed 3D annotations in complex sensor setups.

        

    \subsubsection{Multi-object tracking}
    
            
            
        Multi-object tracking (MOT) aims to estimate how object states (e.g., position, velocity and heading) evolve over time. This temporal information is essential for understanding the intent of dynamic agents and predicting their future motion, which is crucial for planning and decision-making \cite{yurtsever2020survey}.
        Most modern approaches follow the tracking-by-detection paradigm \cite{weng20203d, chiu2020probabilistic3dmultiobjecttracking, yin2021center, kim2021eagermot}. Given detections from previous time steps, such methods aim to make associations across time to create tracks, estimate the state of each object based on the sequence of associated detections, and manage track initialization and termination as objects enter or leave the surveillance area.
        
        One family of methods addresses these aspects under a probabilistic framework based on Random Finite Sets \cite{williams2015marginal, garcia2018poisson}. This approach provides a principled way to model uncertainties through motion and measurement models with recursive Bayesian updates. However, high computational complexity prohibits exact solutions, limiting performance. 
        In practice, most methods instead rely on heuristics or learning-based methods to address parts of the tracking pipeline. For example, in scoring possible associations, measurement likelihood have been replaced by various affinity metrics, based on e.g., intersection-over-union (IoU) \cite{weng20203d}, distance \cite{yin2021center, chiu2020probabilistic3dmultiobjecttracking, teepe2024earlybird}, or appearance features \cite{chiu2021probabilistic, teepe2024earlybird}. Global optimal matching may then be computed with the Hungarian algorithm \cite{weng20203d, teepe2024earlybird}, but greedy methods are also common \cite{yin2021center, chiu2020probabilistic3dmultiobjecttracking}. Moreover, initialization and termination of tracks are often based on simple count-based rules \cite{yin2021center, chiu2020probabilistic3dmultiobjecttracking, weng20203d}.

        Tracking-by-detection has been extensively studied in camera-based surveillance \cite{xu2016multi, cheng2023rest, ong2020bayesian, teepe2024earlybird}, and in LiDAR or camera-based autonomous driving \cite{weng20203d, chiu2020probabilistic3dmultiobjecttracking, yin2021center}.
        Recent methods go beyond pure detection-based tracking: \cite{zhou2020tracking, zhang2023motiontrack} perform joint detection and tracking, while \cite{meinhardt2022trackformer} introduces a transformer-based tracking-by-attention paradigm.
        Recent methods also study V2X-enabled tracking. \cite{chiu2024probabilistic} propose sharing detections between vehicles to enhance subsequent tracking-by-detection robustness. \cite{hu2024integrated} instead fuse features from an infrastructure and onboard LiDAR and perform transformer-based joint detection and tracking. Designing high-performing multi-object tracking methods that are robust and computationally tractable remains an open research challenge, especially in the emerging field of V2X perception.

    \subsubsection{Motion prediction} 

        %
        
        Motion predictions aims to estimate the future location of relevant dynamic objects based on the detection and tracking results. Historical motion profiles of the target and nearby agents, along with environmental layout, are commonly used. Traditional methods use established motion models of dynamic obstacles for prediction, such as the constant velocity model \cite{cvm_2020}, the social force model \cite{sf_1995}, and the velocity obstacle model \cite{rvo_2008}. However, traditional methods exhibit notable limitations. They struggle to account for complex interactions and maneuvers of target objects without a predefined rule base, and face difficulties in generating accurate predictions in complex situations. 
        
        Advancements in deep learning have enabled neural networks to model environmental context and agent interactions while accounting for uncertainty, making them well-suited for dynamic environments. 
        Social pooling \cite{socialgan_2018_Gupta} is a technique that captures interactions among agents by aggregating their features to infer collective behavior, allowing the network to learn motion patterns while explicitly considering these interactions. CNNs are widely used to extract spatial information from the environment \cite{overcoming_2019_Makansi, ynet_2021_Mangalam, ze_2025, fang_2020_tpnet}, thereby enhancing the network’s ability to perceive the layout of the workspace. This is crucial to make adaptive predictions when the surroundings are dynamic and the layout may change over time.
        Recently, the transformer attention mechanism \cite{Trajectronpp_2020_Salzmann, selfattention_2024_yang} and inverse reinforcement learning \cite{end2end_2022_guo} have been used to capture the intricate dependencies between agents and their environments. Still, generating predictions that are accurate and adequately describe the (multi-modal) uncertainty about the future motion of objects remain difficult in complex environments.


    \subsection{Fleet management and planning}


        In the autonomous driving community, deciding the next action for the vehicle is typically accomplished with hierarchical pipeline consisting of global route planning, behavioral decision making, local motion planing, and control \cite{badue2021self, paden2016survey, teng2023motion, zhao2024autonomous}. Similar ideas can naturally be applied to the control of AMRs, although, we are considering an entire fleet rather than a single vehicle. 
        The rationale behind the hierarchical decomposition of the problem is two-fold. First, the uncertainty about the environment grows rapidly with time, which makes it feasible to explicitly consider possible futures only over short time horizons. Second, details that are central to short-term planning, e.g., exact vehicle dynamics and predicted motion of nearby vehicles, may be disregarded in long-term planning as they are unlikely to effect the results.

        
        

        \begin{figure}[h!]
            \centering
            \includegraphics[width=1.0\linewidth]{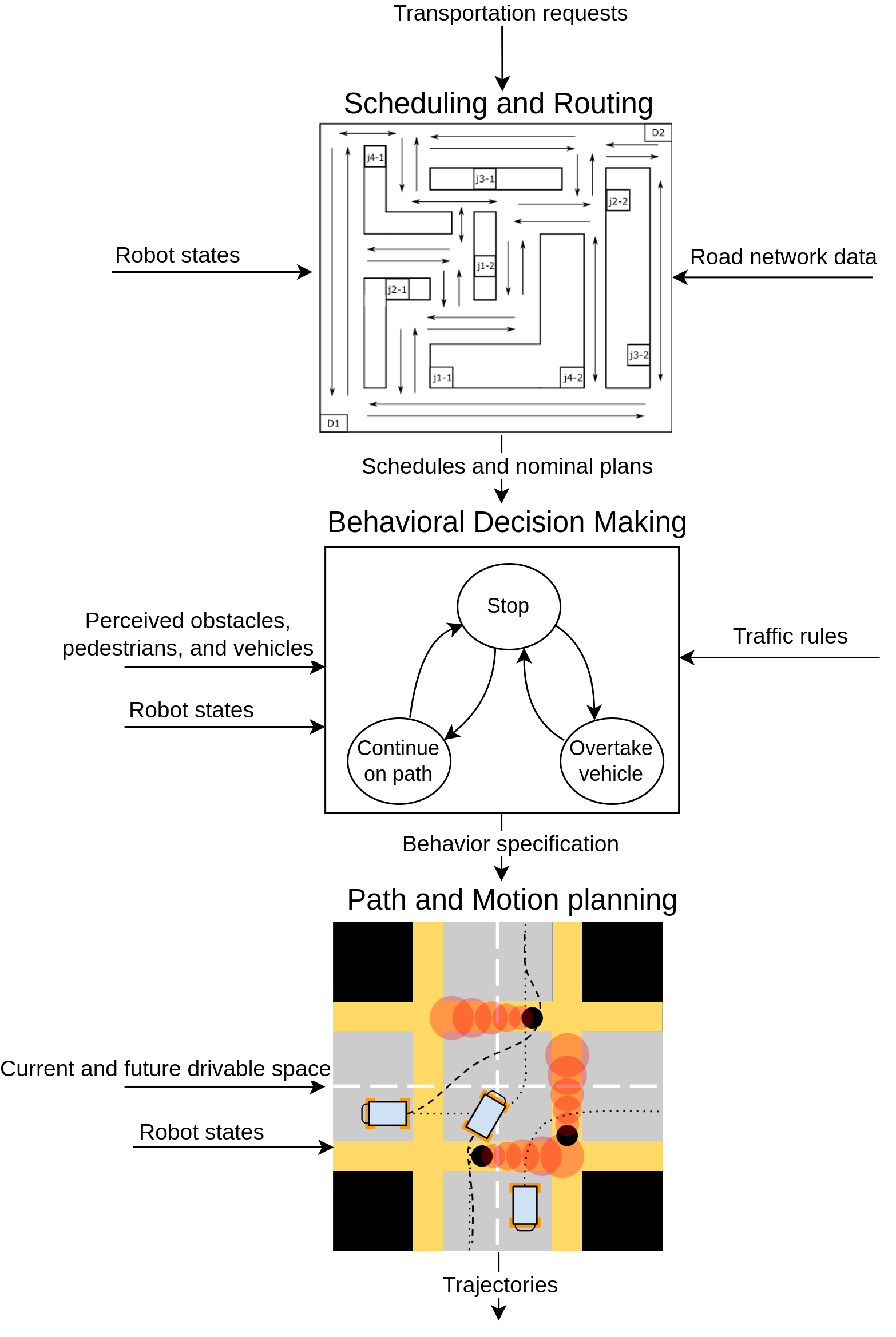}
            \caption{Overview of the fleet management and planning framework. Arrows from the left correspond to output from the perception module, while arrows from the right are user specifications.}
            \label{fig:ctrl-overview}
        \end{figure}

        Figure \ref{fig:ctrl-overview} illustrates a hierarchical planning pipeline for AMRs.
        At the highest level, scheduling and route planning assign tasks to the robots based on the road network and current transportation demands. The considered time horizon for this problem is typically hours. Therefore, only nominal plans can be provided by this layer, which is then supplemented with real-time decision making at lower levels. Behavioral decision making is responsible for determining the course of action over a shorter time horizon, typically not longer than a few minutes. The sequence of action is based on current observations of the environment, e.g., overtake the preceding vehicle, continue on route to the depot station, and come to a stop.
        The selected behavior is often manifested in a reference path provided by a path planner. As available paths are continuously changing in a dynamic environment, path planning is a central component in the pipeline. Subsequently, the motion planning layer computes a trajectory that progresses the robot along the reference path in order to achieve the selected behavior. The trajectory typically spans only a few seconds, which allows for explicit reasoning about possible future states of nearby dynamic objects to produce feasible trajectories.
        Finally, a local feedback controller is employed to follow such trajectory.


        In the following, we review relevant literature on scheduling and route planning, behavioral decision making, path planning, and motion planning. The low-level control is excluded since it is not specific for our application and has already been covered by a large body of literature. It is noteworthy that the distinction between the different layers is somewhat vague, and methods that treat, for example, behavioral decision making and motion planning jointly are common.  

        \subsubsection{Scheduling and route planning}

            Scheduling and routing is typically formulated as a Vehicle Routing Problem (VRP), a class of NP-hard combinatorial optimization problems. Optimization is performed over a graph derived from the predefined road network. The task is to determine which edges each robot should traverse, and at which time, to fulfill transportation jobs. The edge weights typically correspond to distance or estimated travel time. 
            The graph is often directed to account for one-way and two-way roads, and may include features such as edge capacity to manage congestion and depot stations that represent robot start, end, or charging locations. Time windows can also be incorporated to ensure that robots are at specific locations at designated times, which is a common requirement in industrial applications \cite{konstantakopoulos2022vehicle}.
            Commonly, the optimization objective is a function of cumulative traveling distance, tardiness, or level of congestion \cite{lin2022review}. 

            Due to the high computational complexity, solving the scheduling and routing problem simultaneously in a single, monolithic optimization formulation is only practical for relatively small instances. Feasible problems sizes are limited to a few robots and tasks, and a graph with a few dozen nodes. In such cases, both approximate methods \cite{jun2022scheduling} and exact algorithms \cite{roselli2021solving} can be used effectively. 
            For larger systems, scheduling and routing may be treated separately to reduce the computational complexity. Specifically, iterating between first solving the routing problem and then the scheduling problem is a promising approach \cite{roselli2022compositional}. 
            Alternatively, routes and schedules can be computed sequentially for one AMR at the time \cite{popolizio2024online}. While these approaches allow for faster computation and dealing with larger problem instances, they can't guarantee global optimal solutions.
            Another approach to compute routes and schedules is to use machine learning (ML) algorithms \cite{bogyrbayeva2024machine}; the most investigated method, as well as the most promising, is to train reinforcement learning agents end-to-end \cite{silva2019reinforcement,zhang2020multi} to either drive a single AMR, or the entire fleet. 
            On the other hand, ML has also been used in combination with analytical methods \cite{bai2023analytics}; in the modeling phase, ML can be used to estimate model parameters related to features of the problem from historical data, while in the solving phase, ML can be used to quickly generate initial solutions that can be further improved by non-learning methods.



        \subsubsection{Behavioral decision making}

            To account for dynamic elements in the environment, such as nearby traffic and pedestrians, an online planner is required. Typically, planning is based on a domain model that includes possible environmental configurations (states), action definitions with preconditions and effects, and the goal specification \cite{ghallab2004automated}. By searching the state and action space of the model, the planner finds the sequence of actions needed to reach the next scheduled goal.

            In classical planning, the Planning Domain Definition Language PDDL offers a standard for modeling. PDDL defines types, predicates, and actions \cite{mcdermott1998pddl}, and is widely used for structured problem representations. It also has extensions that support temporal aspects of planning \cite{fox2003pddl21}. For reactive systems, wherein asynchronous events ought to trigger certain responses, event-driven formalisms such as Event-Condition-Action (ECA rules) \cite{paton1999active} or Extended Finite Automata (EFAs) \cite{skoldstam2008supervisory} are often more suitable.  

            Planning techniques vary depending on the complexity of the problem. For smaller problems, explicit state-space search methods such as Breadth-First Search are common \cite{russell2016artificial}. For larger state spaces, symbolic methods based on Binary Decision Diagrams (BDDs) \cite{bryant1986graph} or SAT-based planning, which leverages recent advancements in SAT solvers by casting the problem as Boolean Satisfiability \cite{kautz1992planning}, are more effective. To further manage complexity, specifications like Linear Temporal Logic (LTL) can be used to define desired properties over execution paths \cite{pnueli1977temporal}. This helps prune the search space by eliminating solutions that violate critical constraints \cite{baier2008principles}.

            A key remaining challenge is creating the domain model. Manual definition is time-consuming and error-prone, especially in complex or dynamic environments \cite{amir2008learning}. Recent work attempts to learn such models through interaction with a simulator or the real-world system using techniques such as active learning \cite{ashfaq2022learning}. However, reliably learning accurate and comprehensive models remains an open research problem.

        \subsubsection{Path planning}
          
            The path planning problem concerns finding either the shortest or a collision-free path through an environment. A path $\mathcal{P}$ can be defined as an ordered set of connected points or a spline in space, containing an initial point $\mathcal{P}_{\text{start}}$ and a final point $\mathcal{P}_{\text{end}}$. Path planning approaches are commonly categorized as graph-based, sampling-based, or numerical optimization methods \cite{pathplanning2024review}.


            Graph-based methods like Dijkstra's and A* require discretization of the environment into a graph-like structure, and is widely used in static environments \cite{pathplanning2024review}. However, when the environment changes, algorithms such as A* must typically recompute the entire path, which limits their efficiency in dynamic settings \cite{uav2025survey}. D* and its derivatives \cite{d_star_improved} address this limitation by updating only the affected portions of the path when changes are detected, significantly reducing the computational burden in dynamic or partially known environments.

            Sampling-based methods such as Probabilistic Roadmap and Rapidly-Exploring Random Trees \cite{amr_pathplanning} operate directly in continuous configuration spaces by randomly sampling states and incrementally building connectivity graphs or trees. These methods are particularly effective in high-dimensional planning problems due to their ability to explore large state spaces efficiently \cite{2023_dong}, and offer probabilistic completeness despite generally lacking guarantees of optimality.

            Numerical optimization methods such as Ant Colony Optimization (ACO) and Genetic Algorithms (GA) offer flexible, heuristic-driven approaches to path planning. These methods typically require discretization of the search space, for example by modeling the environment as a grid or waypoint graph. ACO uses pheromone-based exploration over such discrete representations to discover globally efficient paths, with recent variants improving convergence speed and robustness in dynamic environments. GA evolves path solutions via fitness-based selection, often optimizing both path length and smoothness within discretized configurations \cite{amr_pathplanning}. Although careful parameter tuning is required and these methods are less suited for strict real-time use, they are effective in offline planning frameworks.

        \subsubsection{Motion planning}

            Collision-free motion planning, also referred to as trajectory planning or local path planning, computes a dynamically feasible trajectory that safely guides a robot from its current state to a desired goal. The trajectory must respect the robot’s kinematic and dynamic constraints, avoid both static and dynamic obstacles, optimize cost terms such as energy consumption, safety margins, execution time, and trajectory smoothness, and adhere to a reference path provided by a global planner \cite{trajectory2025review}. 
            As shown in the bottom part of Fig.~\ref{fig:ctrl-overview}, three AMRs navigate a shared intersection alongside pedestrians, following reference paths (dotted lines) while generating locally feasible trajectories (dashed lines) that react to predicted obstacle motion. 
            Effective planning in such dynamic environments depends both on the chosen algorithm and how obstacles are represented. The translucent red ellipses in the figure depict a probabilistic obstacle model that captures both position and uncertainty over time. Obstacle models may be parametric (e.g., geometric primitives like ellipses and bounding boxes) or non-parametric (e.g., occupancy grids, Euclidean Signed Distance Fields, or Truncated Signed Distance Fields), and either deterministic or probabilistic depending on how uncertainty is treated \cite{trajectory2025review, safe2024survey, uav2025survey}.            
            Motion planning methods are commonly categorized into four classes: reactive, optimization-based, learning-based, and hybrid methods, based on how they process information and compute actions.

            Reactive motion planning methods compute control commands or trajectories in response to current state and local sensor inputs, without relying on long-horizon predictions. These approaches are typically fast and well-suited for dynamic environments. Notable examples include the Dynamic Window Approach (DWA), which searches a window of admissible velocities to select the one that maximizes an objective function while ensuring safe stopping distances, and the Timed Elastic Band, which optimizes a time-parameterized trajectory using soft constraints for obstacle avoidance \cite{trajectory2025review, teb_improved}.

            Optimization-based approaches formulate motion planning as a constrained optimization problem, where the objective is to compute a feasible and cost-efficient trajectory. A prominent example is Model Predictive Control (MPC), which uses a predictive model of the robot's dynamics to iteratively optimize control inputs at each time step over a finite time horizon. MPC incorporates motion constraints such as dynamics, input bounds, and safety requirements, and evaluates a cost function that typically balances tracking performance, smoothness, and control effort. The receding horizon structure makes MPC suited for dynamic and uncertain environments \cite{safe2024survey}. However, solving nonlinear optimization problems in real time imposes a significant computational burden. To alleviate this, many implementations leverage numerical approximations, such as Sequential Quadratic Programming (SQP), or warm-start techniques using precomputed feasible trajectories \cite{safe2024survey}.

            Learning-based motion planning includes both reinforcement learning (RL) and imitation learning (IL). RL learns a policy through trial-and-error interactions with the environment, often in simulation, optimizing cumulative reward signals that encode motion objectives and constraints. Imitation learning, on the other hand, derives policies by mimicking expert demonstrations. These methods are data-intensive and typically require extensive training, but offer good adaptability to dynamic and uncertain environments \cite{intelligent2022review}.

            Hybrid approaches combine different planning paradigms to offset the limitations of individual methods. For example, integrating the global optimality of MPC with the real-time reactivity of RL improves robustness in uncertain environments \cite{mpc-rl, mpc-dqn}. Similarly, combining DWA with RL enables reactive safety while preserving learned behavior \cite{dwa-rl}.

            

            Despite recent progress, motion planning for AMRs remains challenging, particularly when accounting for the robot's kinematics and dynamics alongside timing and collision constraints. Early efforts in this area include trajectory planning based on Signal Temporal Logic (STL) specifications, combined with control strategies such as model predictive control \cite{robusttemporalbelta2015, Charitidoudmpc2024} or control barrier functions \cite{lindemanncbf2019}.
            
            Ensuring real-time feasibility under dynamic and nonlinear constraints remains difficult, especially in unstructured environments \cite{trajectory2024survey}. Uncertainty from imperfect sensing and prediction complicates safety assurance, with deterministic methods often being overly conservative and probabilistic ones lacking formal guarantees \cite{trajectory2024survey}. Scalability is another issue, as high-dimensional or multi-agent scenarios strain most planning algorithms \cite{uav2025survey}. While learning-based methods promise adaptability, they require extensive data, struggle to generalize, and remain hard to interpret or verify \cite{intelligent2022review}. Bridging classical and learning-based approaches while ensuring robustness and safety is a key direction for future research.


            
\section{Industrial evaluation}

We have implemented and deployed a variant of the described system in an industrial setting at Volvo Group Truck Operations (AB Volvo). In this deployment, six robots are used for transporting mufflers from a pre-assembly area to the assembly line. The cycle time is approximately 7 minutes and up to 130 transport operations are performed per day. Each transport task involved a 150-meter drive through a mixed-traffic environment shared with manually operated vehicles and pedestrians. This necessitated robot behaviors capable of handling dynamic and uncertain conditions.

\begin{figure*}[b!]
        \centering
        \includegraphics[width=1\linewidth]{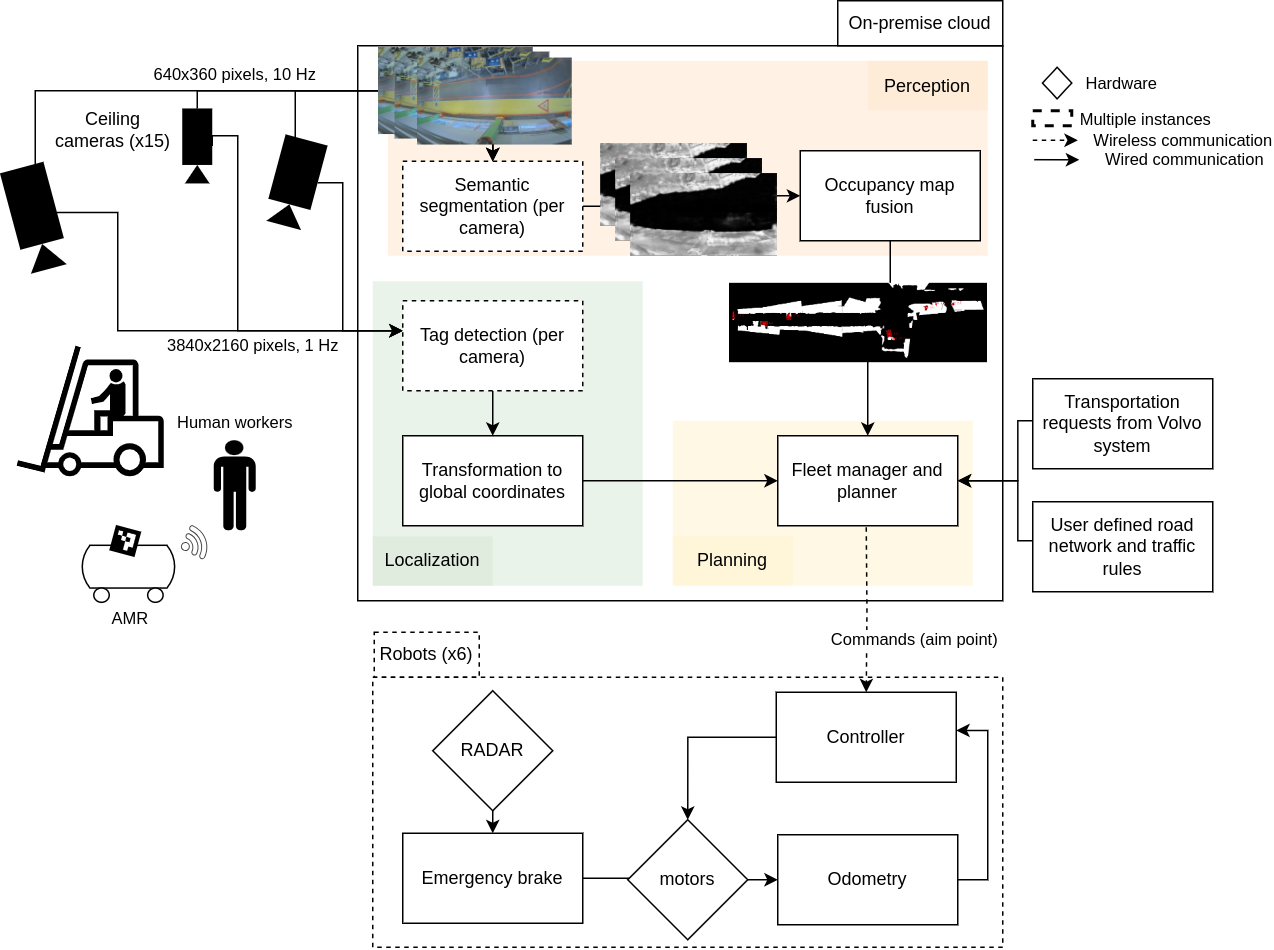}
        \caption{System architecture of the industrial implementation at Volvo.}
        \label{fig:implementation}
    \end{figure*}

The core design principle was to minimize the complexity and cost of the robots by removing expensive onboard sensors and computational units. To this end, localization, perception, and planning capabilities were offloaded to a centralized infrastructure consisting of a local compute cluster and ceiling-mounted cameras. These cameras jointly provide a top-down view of the entire robot-operating area and were used for both robot localization and environmental perception.
Figure~\ref{fig:implementation} illustrates the system architecture. The infrastructure cameras send image streams over Ethernet to the local compute cluster. These images are processed to localize the robots and to build an occupancy map of the environment. Based on the current robot positions, the occupancy map, transportation requests, and a pre-defined road network, a fleet manager assigns tasks and nominal paths to the robots. A path planner then generates motion plans, which are transmitted via Wi-Fi to the robots. The robots execute these plans using low-level controllers, relying on motor odometry as feedback. For safety, an onboard short-range RADAR and a bumper stop activate emergency braking.
The following subsections describe the hardware and software architecture of the system in more detail.

\subsection{Hardware and middle-ware}

    \subsubsection{Robots}

Figure \ref{fig:robot} shows the robot platform, including:
    \begin{itemize}
        \item A safety radar and a bumper stop for collision protection,
        \item     A 48V battery,
        \item     Two motors controlled via a CAN interface,
        \item     Two ESP32-C3 microcontrollers for control and communication,
        \item     A display, lamps, and control buttons,
        \item     A unique ArUco marker \cite{garrido2016generation} (4×4 format, 100 mm size) for localization via the ceiling cameras.
    \end{itemize}
    \begin{figure}[h!]
        \centering
        \includegraphics[width=1\linewidth]{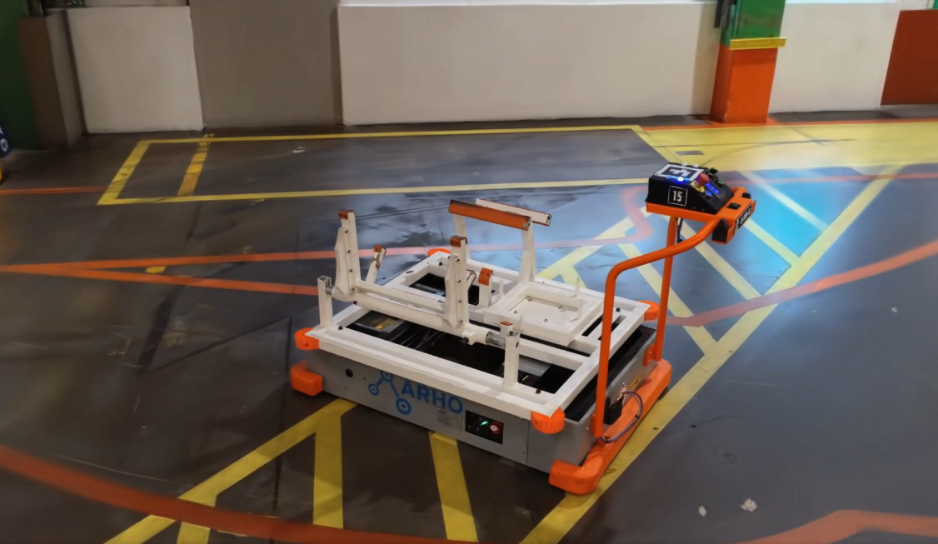}
        \caption{The transportation robot. The scene was reconstructed using Gaussian splatting \cite{ahmed2025realistisk}.}
        \label{fig:robot}
    \end{figure}

 Notably, the robots are not equipped with LiDAR or cameras, which are commonly used for onboard perception and SLAM. Instead, localization and obstacle detection are handled entirely by the ceiling-mounted cameras. Robots receive navigation instructions via Wi-Fi, in the form of simple motion commands and waypoints. The robot firmware is written entirely in Rust, using the async embassy framework.
The robots can also be operated manually. A button on the handle switches the robot to manual mode, allowing human operators to move it as needed. When switched back to autonomous mode, the robot automatically returns to its assigned path.


    \subsubsection{Infrastructure cameras}
\begin{figure*}[h!]
    \centering
    \includegraphics[width=1\linewidth]{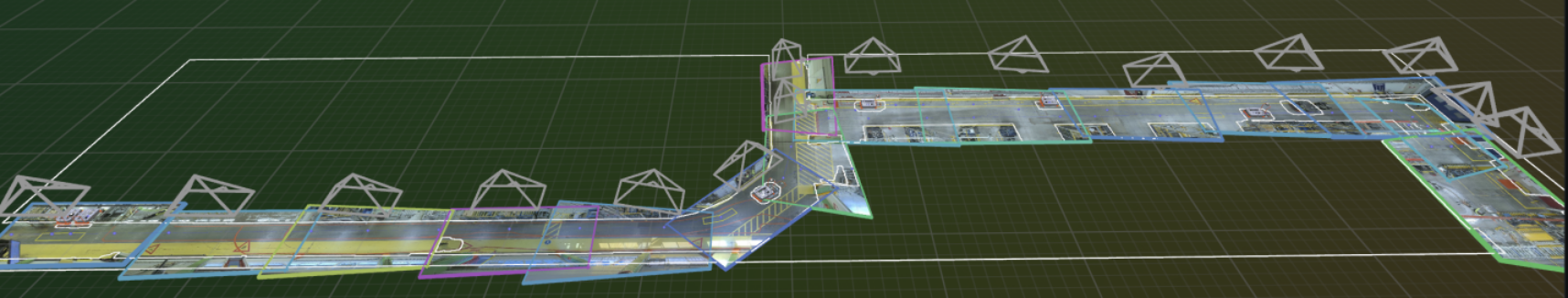}
    \caption{Rerun \cite{RerunSDK} visualization of the deployment area and camera positions. Each of the 15 cameras cover an area of roughly 12$\times$6 meters.}
    \label{fig:camera_layout}
    \end{figure*}

    To enable localization and perception, Power-over-Ethernet (PoE) IP cameras were installed in the factory ceiling. In total, fifteen cameras where installed, which together cover the entire robot operating area. Their layout is shown in Figure \ref{fig:camera_layout}. Most cameras were mounted at a height of approximately 8 meters and oriented in a top-down configuration, each covering about 60 square meters of floor area.
    The cameras were calibrated using the OpenCV toolkit, based on the pinhole camera model and the method described in~\cite{888718}, with a custom-made $2\times1$ meter calibration target. During operation, each camera provides two image streams to the compute cluster:
    \begin{itemize}
        \item Low-resolution (640×360 px) images at 10 Hz, used for obstacle detection;
        \item High-resolution (3840×2160 px) images at 1 Hz, used for robot localization.
    \end{itemize}
    The cameras are not hardware-synchronized, which means that the images from different cameras are captured with slight time difference.

    \subsubsection{Compute cluster}
    A kubernetes cluster was used installed on simple bare metal nodes including one Nvidia A4000 GPU each. The semantic segmentation inference can handles up to 25 cameras per GPU at 10 fps per camera. 
    Most of the code is implemented using rust, making it resource efficient and stable. The cluster is running in an air-tight factory environment in the normal factory network.

    The code structure is simple, using a redis data base as a shared state and the robots communicates with the cluster via the factory wifi network using a simple udp-protocol.

\subsection{Software}
\subsubsection{Localization}
Localization is based on detecting the unique ArUco marker attached to each robot. The high-resolution images from the infrastructure cameras are processed using the OpenCV ArUco library \cite{garrido2016generation, romero2018speeded} to detect these markers. Given that the markers lie on a known horizontal plane (parallel to the factory floor), and at a fixed height, the 2D image coordinates of the marker corners can be mapped to 3D coordinates using the established camera calibration. From these corner positions, the robot’s Cartesian coordinates $(x, y)$ and heading $\theta$ are computed directly. Due to the high resolution of the images, detecting the markers is computationally expensive and is therefore performed at a low frequency (1 Hz). However, onboard the robot, the low-rate position estimates are fused with motor odometry, which enables accurate real-time localization.

\subsubsection{Perception}      

The factory floor is discretized into a uniform grid, with each cell representing a $5\times5$ cm area. The perception system classifies each cell as free or occupied based on semantic segmentation.
Specifically, low-resolution images from each camera are processed using a binary semantic segmentation model to distinguish obstacles from free space. These 2D pixel maps are transformed to the ground plane using a pre-computed homography based on the established camera calibration. Since this homography merely maps the image plane to the ground plane, it disregards the true depth of each pixel in the image. However, since the cameras are mounted at a considerable height, the varying height of the objects in the scene only affects the results modestly. Once the per-camera occupancy maps have been computed, they are fused into a global map. Where cameras have overlapping views, the global map is based on only one of the cameras (typically the closest one). Note that since the cameras are not hardware-synchronized, some errors arise due to time shifts between the different cameras. However, the impact is small given the high processing rate (10 Hz).



\subsubsection{Fleet management and planning}
The deployed system supports a single transportation task type: transporting mufflers from a pickup station to the assembly line and returning. Scheduling is handled using a simple queue-based mechanism. At the start of each day, the three robots queue at the depot. When a new transport request arrives, the next available robot is dispatched.
Execution of each task follows a predefined finite-state machine (FSM), with states and transitions manually defined based on application-specific conditions (e.g., is the robot kitted? is the path clear? are there other vehicles in the intersection?). These checks rely on the perception system and logical evaluations.
Navigation is based on the A-star algorithm, which computes paths over a manually defined road network. Robots follow these paths using a simple motion controller that ``aims" for a point several meters ahead, adjusting speed via ramp profiles based on distance. Heading is regulated via a feedback controller using heading error.

In summary, the robot behavior is intentionally kept simple. Robots stop or slow down when obstacles block their path and yield to higher-priority vehicles. They do not re-plan around obstacles or leave their assigned route. As a result, factory workers must keep robot paths clear for smooth operation. However, the robot may be manually moved off its path, in which case it will automatically navigate back to the assigned route when reactivated.

\subsection{User experience evaluation}

As mobile and collaborative robots are integrated into industrial settings, there is a growing need for human-centered design and interaction \cite{alenljung2023towards}. This evolution underscores the significance of user experience (UX) in human-robot collaboration, focusing on trust, comfort, and communication alongside traditional concerns for safety and efficiency \cite{lindblom2020anemone}. Key UX challenges faced by workers in shared workspaces with mobile robots were investigated in \cite{alenljung2025introducing} to inform human-centered design strategies that enhance safety, productivity, and well-being.
A post-evaluation was conducted two months after the AMR field test, using the UX Questionnaire \cite{ueq2024} and open-ended interviews with eight participants (four operators and four truck drivers) who worked closely with the robots. The UX evaluation covered six aspects: effectiveness, efficiency, safety, sustainability, pleasure, and predictability \cite{alenljung2025introducing}.

Effectiveness was partly satisfactory. Operators found the robot helpful for material delivery and reducing manual tasks, while truck drivers felt it added complexity and delays. Efficiency was rated as partly satisfying; operators appreciated its ease of use and time savings, but truck drivers struggled with unclear behavior, detection issues, and lack of training. Both groups needed better handling of unexpected situations and more support during onboarding. Safety was unsatisfactory, especially for truck drivers facing visibility issues, collision risks, and unclear statuses. Operators generally felt safe but were concerned about the robot’s ability to detect obstacles, such as forklifts. Sustainability ratings were low due to cognitive strain on truck drivers and worries about long-term conditions. Operators found it slightly acceptable but noted ergonomic deficiencies and the need for constant monitoring. Despite some frustrations, pleasure received the highest rating, as both groups found the robots interesting and innovative. Predictability feedback was mixed; operators found the robots largely predictable, while truck drivers encountered unpredictable behavior in non-standard scenarios.

Key issues included challenges in both the physical and cognitive work environments. Physical problems encompassed difficulties in detecting the robot and collision risks. Challenges in the cognitive work environment, such as unclear status, lack of training, increased task demands, workflow disruptions, authority ambiguities, and unclear benefits, further obstructed acceptance. Concerns were also raised regarding the scalability of the system and potential negative impacts on truck drivers’ future roles. Overall, while the robot offers innovation and some operational support, improvements are needed to fully assist truck drivers and operators \cite{alenljung2025introducing}.

The evaluation revealed varied user experiences: operators reported better UX than truck drivers, who faced more issues. Both groups acknowledged company-level benefits but saw few personal gains, highlighting the need for human-centered robot design \cite{alenljung2025introducing}.
Key recommendations include enhancing robot visibility, providing status feedback, offering user-specific training, and establishing clear traffic management rules to accommodate diverse work conditions and user needs \cite{alenljung2025introducing}. This evaluation emphasizes the crucial need to address the varying experiences of operators and truck drivers through UX-driven design, ensuring that human well-being, trust, and effective collaboration with robots are prioritized \cite{alenljung2023towards, lindblom2020anemone}.
\section{Challenges and Future Outlook}
To meet the industry's growing demands for flexibility, autonomous mobile robots (AMRs) must operate safely and efficiently in dynamic and uncertain environments, often alongside human workers. Infrastructure-based sensing and computing provide a promising path forward by extending perception, increasing computational capacity, and enabling global optimization across fleets. However, fully realizing the benefits of such systems requires advances across a wide range of technical domains. 
In what follows, we group the key challenges into four thematic areas: core autonomy and intelligence, infrastructure challenges, system-level considerations, and human-centered design. For each area, we outline open problems and future research directions.

\subsection{Core Autonomy and Intelligence}
\subsubsection{Localization}
        Localization is widely recognized as a key challenge for AMRs, particularly in dynamic and unstructured environments. Infrastructure-based methods, such as those relying on ceiling-mounted cameras, can offer high localization accuracy and are relatively straightforward to implement in controlled settings. 
        However, issues such as occlusions, sensor faults, or limited coverage can disrupt such localization methods, making fallback strategies like onboard SLAM essential. Meanwhile, onboard SLAM faces its own challenges. Performance typically degrades in environments that are highly dynamic, cluttered, or subject to frequent layout changes. As a result, infrastructure-sparse settings remain difficult for current systems, and advancing localization in such contexts is an important direction for future research.

        A promising direction is to combine infrastructure-based and onboard localization. This includes collaborative and cloud-based SLAM methods and the integration of dedicated localization technologies such as Ultra-Wideband (UWB), RFID, Wi-Fi, Bluetooth, Zigbee, artificial landmarks, and infrastructure-mounted cameras. However, fusing these modalities introduces new technical challenges in data fusion, temporal and spatial synchronization, and real-time communication. Addressing these will be essential for reliable, scalable localization in industrial AMR deployments.

\subsubsection{Perception}
        Perception in autonomous systems largely depends on deep learning models for tasks such as feature extraction and object detection. These models are typically trained in a supervised manner, making their performance highly dependent on the availability of large labeled datasets, which are time-consuming and costly to create. A key challenge, therefore, lies in developing generalizable perception models that can perform well with limited labeled data. Approaches such as semi-supervised learning and unsupervised domain adaptation aim to address this by leveraging small labeled datasets alongside larger collections of unlabeled data. More recently, foundation vision models have shown promise in mitigating these limitations by providing universal feature extractors that can simplify various downstream tasks.     

        Additional challenges arise when considering the specific application, involving on-premise cloud processing of heterogeneous data streams from both infrastructure and onboard sensors. Such systems must handle spatial and temporal misalignment caused by calibration errors and communication delays. Moreover, due to limited wireless bandwidth, it is often impractical to transmit raw sensor data to the cloud. Instead, perception algorithms must be designed to operate under communication constraints and prioritize information that contributes most effectively to the system’s performance. 

        Another critical challenge is the quantification of uncertainty, which is essential for robust planning and decision-making. Aleatoric uncertainty arises from inherent factors such as sensor noise and occlusions in the environment. Epistemic uncertainty, on the other hand, stems from model limitations, such as insufficient training data or poor generalization to novel scenarios. Predicting the future state of dynamic environments introduces additional uncertainty, particularly when human behavior must be anticipated. These uncertainties are especially difficult to model when fusing data from multiple modalities and perspectives, as their interactions can affect the overall confidence in non-trivial ways.  
        
        More broadly, ensuring robust perception in dynamic environments with frequent occlusions and unpredictable events remains a major challenge. The perception pipeline typically includes tightly coupled components such as detection, tracking, and motion prediction. Errors in early stages can propagate and compromise overall system performance. Improving robustness across this interdependent pipeline is an active and ongoing area of research.

\subsubsection{Fleet Management and Planning}
        Challenges in fleet management and planning span the entire decision-making stack of AMRs.
        At the high level, scheduling and route planning must cope with growing computational complexity as fleet sizes and routing alternatives increase. These plans also need to remain adaptable to unforeseen disturbances. Ongoing research focuses on developing scalable algorithms capable of efficient rescheduling and robust plan repair in response to disruptions.
        At the mid level, behavioral decision making depends on accurate and comprehensive models of the environment and robot behavior. Specifying these models manually is time-consuming and error-prone, while learning them from data presents its own set of challenges.
        At the low level, path and motion planning in dynamic, unpredictable environments remains difficult. Challenges include computational efficiency, safety guarantees, and handling uncertain perception data.
        
        Beyond these layer-specific issues, the hierarchical structure itself introduces fundamental limitations. Decisions at each level are often made in isolation, based on abstracted and locally available information, without full awareness of constraints or uncertainties encountered at lower levels. While this architecture supports modularity and computational tractability, it can lead to suboptimal or fragile behavior, especially in uncertain environments where robustness is critical.
        The abstraction process tends to omit essential context, such as real-time disturbances, resource contention, or spatio-temporal variability. This hinders the higher layers’ ability to anticipate whether their plans will remain feasible or resilient downstream. As each level optimizes its own objectives independently, the compounded effects can yield plans that are neither robust nor valid for the system as a whole. 
        A key challenge lies in dynamically updating these abstracted representations at runtime. For instance, if a route consistently takes longer than expected, the high-level planner must integrate that feedback to avoid repeated failures in future plans.

\subsubsection{Lifecycle Management of ML Components}
        Developing and deploying dependable machine learning (ML) faces significant challenges related to the inherent uncertainty of ML models. These include collecting and preparing representative datasets, mitigating model epistemic uncertainty (resulted from functional insufficiency) and managing aleatoric (irreducible) uncertainty introduced by factors such as annotation errors, sensor noise, and environmental condition variations. Furthermore, unlike traditional software where full test coverage can be achieved via code scanning, scenario-based testing seems to be the only viable approach for ML based systems. However, it requires large and diverse datasets that represent the set of test scenarios, requiring significant data collection and annotation efforts. Recreating hazardous scenarios for data collection can be challenging or sometimes infeasible. Even with enormous data collection investment, achieving sufficient coverage of all possible data variations remains difficult. 
        
        A promising approach is to augment real-world datasets with synthetic data generated by simulation, increasing variations and coverage. A key challenge is to generate diverse and realistic scenarios along with sensor data, such as images and LiDAR scans, which can allow for closed-loop testing of the system. Recent techniques based on Gaussian Splatting \cite{hess2025splatad} and NeRFs \cite{tonderski2024neurad} hold significant potential as they can be used both to reconstruct real-world scenarios and generate realistic sensor data from novel view-points.
        Despite the potential of facilitating data collection and reducing annotation costs drastically, the sim-to-real gap remains a challenge. More research is required to bridge the simulated and real-world domains for perception and decision making tasks alike. This involves improvements in multiple aspects: making the simulation more realistic, and improving model generalization capabilities through model design and training methods. 
        
        Finally, regardless of how well data collection and model engineering have been performed, unknown/unsafe scenarios will inevitably arise during system operation. Therefore, ML based systems must be equipped with supervisory monitoring mechanisms to timely detect unforeseen operational hazardous risks and support continual data collection for future model updates. Unfortunately, this introduces a new challenge: training on new data risks compromising previously certified performance, requiring rigorous and costly verification and validation (V\&V) activities for each model release.

\subsection{Infrastructure Challenges}

\subsubsection{Communication and Networking}
Seamless wireless communication between robots, infrastructure sensors, and edge computing clusters is essential for the safe and efficient operation of infrastructure-based AMRs. Since the system transmits time-critical data, it is highly sensitive to network disruptions and delays. In particular, unpredictable variations in latency, known as jitter, pose significant challenges for perception and planning systems and must be carefully mitigated. Additionally, because AMRs typically move across large factory areas, maintaining consistent wireless coverage and seamless handover between access points is critical. This is especially difficult in dense industrial environments filled with machinery, assembly lines, and vehicles, all of which can cause signal interference.

Technologies such as 5G and Wi-Fi 7 are designed to address these challenges. 5G enables ultra-reliable and low-latency communication (URLLC) through features such as millimeter wave (mmWave) and flexible numerology, making it suitable for real-time robotics applications. Its support for beamforming and massive MIMO can improve signal strength and reliability in complex factory environments. Wi-Fi 7, on the other hand, offers high throughput using wider channels (up to 320 MHz) and 4096-QAM modulation, making it well-suited for transmitting high-bandwidth data such as video. Features like multi-link and multi-AP operation further enhance connection stability as AMRs move across coverage zones.
Although both technologies provide significant benefits, several challenges still need to be addressed. These include optimal base station placement for private 5G networks, compatibility and interference with existing systems for Wi-Fi 7, and the need for dynamic network management to ensure quality of service (QoS) in constantly changing environments. For a more detailed discussion, we refer the reader to \cite{john2024industry}.

\subsubsection{Cloud Computing and Data Processing}
    Because perception and decision-making are offloaded to an on-premise cloud, the system must support real-time data processing, scalability, and fault tolerance. One major challenge is dynamic resource allocation, as compute demands vary depending on robot location and task complexity. In infrastructure-sparse areas, AMRs must rely on onboard sensing and computation. In contrast, infrastructure-rich zones can produce high volumes of data from multiple sources, such as ceiling-mounted cameras and onboard sensors, which may cause temporary spikes in processing requirements. A critical design question concerns which processing tasks should be offloaded to the cloud and how competing requests should be prioritized \cite{saha2018comprehensive}.
    
    By allocating resources efficiently, the footprint of the on-premise cloud can be minimized, which is an important consideration when scaling the system to hundreds of robots and infrastructure devices. However, achieving this efficiency poses significant challenges, particularly in maintaining real-time performance under fluctuating loads. Moreover, the on-premise cloud service must be robust to node failures, software errors, and hardware degradation, which requires redundancy, distributed processing capabilities, and health monitoring mechanisms.

\subsection{System-Level Considerations}

\subsubsection{System Integration}
System integration introduces significant challenges across hardware and software.
On the hardware side, sensor types and placements must be carefully planned, both onboard the robots and in the environment, to ensure adequate coverage and accuracy for localization and perception tasks. Network design is equally important for maintaining low latency and high reliability, particularly in large or frequently changing environments. Factory layouts that evolve over time, for example through the relocation of machinery or storage areas, may require reconfiguration of sensors and networks to preserve system performance. These dynamics introduce significant maintenance challenges.

The hardware design directly shape system-level characteristics like sensing coverage, latency, communication jitter, and computational availability. These characteristics, in turn, place constraints on the software stack. For example, limited sensor visibility or intermittent network delays must be accounted for in perception algorithms, which may need to operate with incomplete or delayed data. Similarly, decision-making components must be designed to tolerate fluctuations in response time and handle transitions between centralized and local execution. Effectively, software modules must be designed to compensate for hardware and network limitations, which adds complexity to their design and integration.

Although often treated as implementation details, these cross-cutting concerns profoundly affect system performance, reliability, and scalability. The intricate dependencies between hardware, software, and communication layers can lead to emergent behaviors that are difficult to predict and diagnose. This complexity highlights the need for principled, systems-level approaches to AMR integration that go beyond modular software design and consider the full stack as a unified whole.


\subsubsection{Safety Assurance}
    Because robots operate alongside unprotected humans and manually driven vehicles, safety assurance is essential. Deep learning (DL) models, which are the core of the system intelligence, are particularly opaque in nature. They often consists of huge neural networks with millions of parameters. Consequently, there are very limited options to analyze misbehavior from a functional safety perspective, as neither traditional code review nor safety assurance practices are applicable. Ongoing research is focused on developing methodologies that enable the safe development and integration of DL-based components into safety-critical systems. The latest developments in the safety assurance guidelines for DL-based systems include AMLAS \cite{hawkins_guidance_2021} and, most recently, AI-FSM \cite{safexplain_d21_2024}, together with experimental compliance practices \cite{borg_ergo_2023, safexplain_d33_2025} and relevant standards \cite{sotif_2022, tr5469}. Although these frameworks offer guidance for adopting ML in safety-critical applications, significant challenges remain in building fully compliant complex systems, including deriving allocated safety metrics to individual DL components. 
\subsubsection{Testing}
Manual testing remains common in industry, but verifying autonomous systems with machine learning-based perception and uncertain actor dynamics poses major challenges. Scenario-based verification is increasingly needed to support safety arguments. Tools like scenarion-based generation \cite{scenic2019} help generate diverse, realistic scenarios, while falsification techniques \cite{donze2010breach,verifaicav19} enables search-based falsification against temporal logic specifications. These methods increase demands on the simulation environment, which must accurately model both the physical dynamics and the perception system, including sensor noise and classification errors. 

\subsubsection{Security and Privacy}
Infrastructure-based AMR systems face significant security challenges, particularly in wireless communication and control system resilience. Wireless communication exposes the system to threats such as jamming, spoofing, and man-in-the-middle attacks \cite{sedar2023comprehensive}. Effective protection requires both proactive measures like secure authentication and reactive strategies such as real-time network anomaly detection. Additionally, AI-based control and perception systems are vulnerable to adversarial inputs that can cause unsafe behavior \cite{girdhar2023cybersecurity}, emphasizing the need for robust models and runtime safeguards.

Privacy risks arise from visual sensors that may capture sensitive data, including images of workers, proprietary products, or confidential processes. Storing such data for AI training raises legal and ethical concerns, particularly under regulations like the GDPR.
An emerging solution is federated learning \cite{zhang2021survey}, where AI models are trained locally at each deployment site using only local data. Instead of sharing raw data, only model updates are sent to a central server. This method helps protect privacy by keeping sensitive data within local boundaries while still enabling system-wide improvements.

\subsection{Human-Centered Design}

\subsubsection{Human-Robot Interaction}
    The user experience evaluation revealed several challenges in human-robot interaction within shared industrial environments, particularly regarding collaboration, communication, and user acceptance. Forklift drivers, in particular, reported a high mental workload when interacting with the AMRs because they found the robots difficult to spot and their behavior occasionally unpredictable. Identifying the specific causes of these negative experiences and determining how to mitigate them remains a complex and open-ended problem. In fact, it aligns with the notion of a wicked problem \cite{rittel1973dilemmas}, where the problem definition itself is fluid and solutions are difficult to evaluate definitively.
        
    One promising way to reduce cognitive strain is to establish clear traffic rules and interaction protocols. These reduce ambiguity, minimize ad hoc decision-making, and help human workers better anticipate robot behavior. However, designing rules that ensure both safety and efficiency while allowing flexibility is challenging. In practice, strict adherence to predefined rules is not always feasible. Unexpected obstacles, robot malfunctions, or deviations from normal operations may require temporary deviations from standard behavior.

    In the deployment at Volvo, humans could manually take control of the robots to handle unexpected situations. While effective, this interrupts workflow and increases the workload for the humans, potentially hindering acceptance of the system.
    Improving collaboration between humans and robots is therefore essential. Robots must be able to infer human intentions through cues such as gestures, gaze, or context, while clearly communicating their own intentions in ways that are intuitive for non-experts. Beyond predictable motion, tools like LED indicators, sounds, or display screens can improve transparency and reduce uncertainty. 
    
    If not carefully designed, however, systems that require users to interpret multiple signals or give nuanced input to the system may inadvertently increase cognitive load further.
    A human-centered design approach must therefore strike a careful balance: enabling expressive, intuitive communication to enable seamless collaboration between human and robots, while preserving simplicity and minimizing the user’s need to learn or adapt to overly complex robot behavior.

\section{Conclusions}

    To address the fragmented landscape of infrastructure-based AMR systems, this paper presents a modular reference architecture that combines infrastructure-based sensing, on-premise cloud computing, and onboard autonomy. We reviewed enabling technologies for localization, perception, and planning; and further demonstrated the practical feasibility of the approach through a real-world industrial deployment. Based on the review and industrial evaluation, we identified challenges in several research areas that warrant further investigation.   
    As robotics moves beyond isolated automation solutions toward more collaborative, scalable, and adaptive systems, we argue that infrastructure-enabled autonomy will play a central role. We hope this paper provides a foundation for both academic research and industrial adoption of the next generation of AMR systems.

\textbf{Acknowledgements} This work was supported by AB Volvo, the Wallenberg AI, Autonomous Systems and Software Program (WASP) funded by the Knut and Alice Wallenberg Foundation, and the Vinnova funded projects SMILE IV (2023-00789) and AIHURO (2022-03012).


{
    \small
    \bibliographystyle{ieeenat_fullname}
    \bibliography{main}

@String(CVPR= {IEEE Conf. Comput. Vis. Pattern Recog.})

@String(ICCV= {IEEE/CVF Int. Conf. Comput. Vis.})

@String(ECCV= {Eur. Conf. Comput. Vis.})

@String(NIPS= {Adv. Neural Inform. Process. Syst.})

@STRING(CASE = {IEEE Int. Conf. Autom. Sci. Eng.})

@String(ICRA = {IEEE Int. Conf. Robot. Autom.})

@String(WACV = {IEEE/CVF Winter Conf. Appl. Comput. Vis.})

@String(ESA = {Expert Syst. Appl.})

@String(IROS = {IEEE/RSJ Int. Conf. Intell. Robots Syst.})

@STRING(JAIR = {J. Artif. Intell. Res.})

@STRING(CAV = {Int. Conf. Comput. Aided Verif.})

@STRING(TPAMI = {IEEE Trans. Pattern Anal. Mach. Intell.})

@STRING(TEV = {IEEE Trans. Intell. Veh.})

@article{yurtsever2020survey,
  title={A survey of autonomous driving: Common practices and emerging technologies},
  author={Yurtsever, Ekim and Lambert, Jacob and Carballo, Alexander and Takeda, Kazuya},
  journal={IEEE access},
  volume={8},
  pages={58443--58469},
  year={2020},
  publisher={IEEE}
}

@article{bernardo2022survey,
  title={Survey on robotic systems for internal logistics},
  author={Bernardo, Rodrigo and Sousa, Jo{\~a}o MC and Gon{\c{c}}alves, Paulo JS},
  journal={J. Manuf. Syst.},
  volume={65},
  pages={339--350},
  year={2022},
  publisher={Elsevier}
}

@article{panigrahi2022localization,
  title={Localization strategies for autonomous mobile robots: A review},
  author={Panigrahi, Prabin Kumar and Bisoy, Sukant Kishoro},
  journal={J. King Saud Univ.–Comput. Inf. Sci.},
  volume={34},
  number={8},
  pages={6019--6039},
  year={2022},
  publisher={Elsevier}
}

@article{niloy2021critical,
  title={Critical design and control issues of indoor autonomous mobile robots: A review},
  author={Niloy, Md AK and Shama, Anika and Chakrabortty, Ripon K and Ryan, Michael J and Badal, Faisal R and Tasneem, Zinat and Ahamed, Md H and Moyeen, Sumaya I and Das, Sajal K and Ali, Md F and others},
  journal={IEEE Access},
  volume={9},
  pages={35338--35370},
  year={2021},
  publisher={IEEE}
}

@article{fragapane2021planning,
  title={Planning and control of autonomous mobile robots for intralogistics: Literature review and research agenda},
  author={Fragapane, Giuseppe and De Koster, Rene and Sgarbossa, Fabio and Strandhagen, Jan Ola},
  journal={Eur. J. Oper. Res.},
  volume={294},
  number={2},
  pages={405--426},
  year={2021},
  publisher={Elsevier}
}

@article{huang2023indoor,
  title={Indoor positioning systems of mobile robots: A review},
  author={Huang, Jiahao and Junginger, Steffen and Liu, Hui and Thurow, Kerstin},
  journal={Robotics},
  volume={12},
  number={2},
  pages={47},
  year={2023},
  publisher={MDPI}
}

@article{john2024industry,
  title={Industry 4.0 and Beyond: The Role of 5G, WiFi 7, and Time-Sensitive Networking (TSN) in Enabling Smart Manufacturing},
  author={John, Jobish and Noor-A-Rahim, Md and Vijayan, Aswathi and Poor, H Vincent and Pesch, Dirk},
  journal={Future Internet},
  volume={16},
  number={9},
  pages={345},
  year={2024},
  publisher={Multidisciplinary Digital Publishing Institute (MDPI)}
}

@inproceedings{tonderski2024neurad,
  title={Neurad: Neural rendering for autonomous driving},
  author={Tonderski, Adam and Lindstr{\"o}m, Carl and Hess, Georg and Ljungbergh, William and Svensson, Lennart and Petersson, Christoffer},
  booktitle=CVPR,
  pages={14895--14904},
  year={2024}
}

@inproceedings{hess2025splatad,
  title={Splatad: Real-time lidar and camera rendering with 3d gaussian splatting for autonomous driving},
  author={Hess, Georg and Lindstr{\"o}m, Carl and Fatemi, Maryam and Petersson, Christoffer and Svensson, Lennart},
  booktitle=CVPR,
  pages={11982--11992},
  year={2025}
}

@article{alam2024fiducial,
  title={Fiducial Markers and Particle Filter Based Localization and Navigation Framework for an Autonomous Mobile Robot},
  author={Alam, Muhammad Shahab and Gullu, Ali Ihsan and Gunes, Ahmet},
  journal={SN Comput. Sci.},
  volume={5},
  number={6},
  pages={748},
  year={2024},
  publisher={Springer}
}

@article{mohanarajah2015cloud,
  title={Cloud-based collaborative 3D mapping in real-time with low-cost robots},
  author={Mohanarajah, Gajamohan and Usenko, Vladyslav and Singh, Mayank and D'Andrea, Raffaello and Waibel, Markus},
  journal={IEEE Trans. Autom. Sci. Eng.},
  volume={12},
  number={2},
  pages={423--431},
  year={2015},
  publisher={IEEE}
}

@inproceedings{balogh2021cloud,
  title={Cloud-controlled autonomous mobile robot platform},
  author={Balogh, Marcell and Vid{\'a}cs, Attila and Feh{\'e}r, G{\'a}bor and Maliosz, Markosz and Horv{\'a}th, M{\'a}rton {\'A}ron and Reider, Norbert and R{\'a}cz, S{\'a}ndor},
  booktitle={IEEE 32nd Annu. Int. Symp. Pers., Indoor Mobile Radio Commun.},
  pages={1--6},
  year={2021},
  organization={IEEE}
}

@inproceedings{zhang2018autonomous,
  title={An autonomous robotic system for intralogistics assisted by distributed smart camera network for navigation},
  author={Zhang, Xu and Scholz, Michael and Reitelsh{\"o}fer, Sebastian and Franke, J{\"o}rg},
  booktitle=CASE,
  pages={1224--1229},
  year={2018},
  organization={IEEE}
}

@article{garrido2016generation,
  title={Generation of fiducial marker dictionaries using mixed integer linear programming},
  author={Garrido-Jurado, Sergio and Mu{\~n}oz-Salinas, Rafael and Madrid-Cuevas, Francisco Jos{\'e} and Medina-Carnicer, Rafael},
  journal={Pattern recognition},
  volume={51},
  pages={481--491},
  year={2016},
  publisher={Elsevier}
}

@article{romero2018speeded,
  title={Speeded up detection of squared fiducial markers},
  author={Romero-Ramirez, Francisco J and Mu{\~n}oz-Salinas, Rafael and Medina-Carnicer, Rafael},
  journal={Image Vis. Comput.},
  volume={76},
  pages={38--47},
  year={2018},
  publisher={Elsevier}
}

@inproceedings{alenljung2025introducing,
  title={Introducing Mobile Robots on the Shop Floor: User Experience Issues},
  author={Alenljung, Beatrice and Lindblom, Jessica},
  booktitle={Int. Conf. Hum.-Comput. Interact.},
  pages={3--19},
  year={2025},
  organization={Springer}
}

@article{rubio2019review,
  title={A review of mobile robots: Concepts, methods, theoretical framework, and applications},
  author={Rubio, Francisco and Valero, Francisco and Llopis-Albert, Carlos},
  journal={Int. J. Adv. Robot. Syst.},
  volume={16},
  number={2},
  pages={1729881419839596},
  year={2019},
  publisher={SAGE Publications Sage UK: London, England}
}

@article{alatise2020review,
  title={A review on challenges of autonomous mobile robot and sensor fusion methods},
  author={Alatise, Mary B and Hancke, Gerhard P},
  journal={IEEE access},
  volume={8},
  pages={39830--39846},
  year={2020},
  publisher={IEEE}
}

@misc{ueq2024,
  author       = {{User Experience Questionnaire (UEQ)}},
  title        = {{UEQ - User Experience Questionnaire}},
  howpublished = {\url{https://www.ueq-online.org/}},
  note         = {Accessed: Jul. 1, 2025},
  year         = {2024}
}

@article{sedar2023comprehensive,
  title={A comprehensive survey of V2X cybersecurity mechanisms and future research paths},
  author={Sedar, Roshan and Kalalas, Charalampos and V{\'a}zquez-Gallego, Francisco and Alonso, Luis and Alonso-Zarate, Jesus},
  journal={IEEE Open J. Commun. Soc.},
  volume={4},
  pages={325--391},
  year={2023},
  publisher={IEEE}
}

@article{girdhar2023cybersecurity,
  title={Cybersecurity of autonomous vehicles: A systematic literature review of adversarial attacks and defense models},
  author={Girdhar, Mansi and Hong, Junho and Moore, John},
  journal={IEEE Open J. Veh. Technol.},
  volume={4},
  pages={417--437},
  year={2023},
  publisher={IEEE}
}

@article{zhang2021survey,
  title={A survey on federated learning},
  author={Zhang, Chen and Xie, Yu and Bai, Hang and Yu, Bin and Li, Weihong and Gao, Yuan},
  journal={Knowl.-Based Syst.},
  volume={216},
  pages={106775},
  year={2021},
  publisher={Elsevier}
}

@article{zhang2023automated,
  title={Automated guided vehicles and autonomous mobile robots for recognition and tracking in civil engineering},
  author={Zhang, Jianqi and Yang, Xu and Wang, Wei and Guan, Jinchao and Ding, Ling and Lee, Vincent CS},
  journal={Autom. Constr.},
  volume={146},
  pages={104699},
  year={2023},
  publisher={Elsevier}
}

@article{fottner2021autonomous,
  title={Autonomous systems in intralogistics: state of the Art and future research challenges},
  author={Fottner, Johannes and Clauer, Dana and Hormes, Fabian and Freitag, Michael and Beinke, Thies and Overmeyer, Ludger and Gottwald, Simon Nicolas and Elbert, Ralf and Sarnow, Tessa and Schmidt, Thorsten and others},
  journal={Logist. Res.},
  volume={14},
  number={1},
  pages={1--41},
  year={2021},
  publisher={Bremen: Bundesvereinigung Logistik (BVL)}
}

@article{lackner2024review,
  title={Review of autonomous mobile robots in intralogistics: state-of-the-art, limitations and research gaps},
  author={Lackner, Thorge and Hermann, Julian and Kuhn, Christian and Palm, Daniel},
  journal={Procedia CIRP},
  volume={130},
  pages={930--935},
  year={2024},
  publisher={Elsevier}
}

@article{lee2003controlling,
  title={Controlling mobile robots in distributed intelligent sensor network},
  author={Lee, Joo-Ho and Hashimoto, Hideki},
  journal={IEEE Trans. Ind. Electron.},
  volume={50},
  number={5},
  pages={890--902},
  year={2003},
  publisher={IEEE}
}

@article{reina2014zeppelin,
  title={zePPeLIN: Distributed path planning using an overhead camera network},
  author={Reina, Andreagiovanni and Gambardella, Luca Maria and Dorigo, Marco and Di Caro, Gianni A},
  journal={Int. J. Adv. Robot. Syst.},
  volume={11},
  number={8},
  pages={119},
  year={2014},
  publisher={SAGE Publications Sage UK: London, England}
}

@inproceedings{streit2016vision,
  title={Vision-based path construction and maintenance for indoor guidance of autonomous ground vehicles based on collaborative smart cameras},
  author={Streit, Franz-Josef and Pantho, Md Jubaer Hossain and Bobda, Christophe and Roullet, Cindy},
  booktitle={Int. Conf. Distrib. Smart Cameras},
  pages={44--49},
  year={2016}
}

@article{hakak2023autonomous,
  title={Autonomous Vehicles in 5G and beyond: A Survey},
  author={Hakak, Saqib and Gadekallu, Thippa Reddy and Maddikunta, Praveen Kumar Reddy and Ramu, Swarna Priya and De Alwis, Chamitha and Liyanage, Madhusanka and others},
  journal={Veh. Commun.},
  volume={39},
  pages={100551},
  year={2023},
  publisher={Elsevier}
}

@article{liu2020computing,
  title={Computing systems for autonomous driving: State of the art and challenges},
  author={Liu, Liangkai and Lu, Sidi and Zhong, Ren and Wu, Baofu and Yao, Yongtao and Zhang, Qingyang and Shi, Weisong},
  journal={IEEE Internet Things J.},
  volume={8},
  number={8},
  pages={6469--6486},
  year={2020},
  publisher={IEEE}
}

@inproceedings{song2024collaborative,
  title={Collaborative semantic occupancy prediction with hybrid feature fusion in connected automated vehicles},
  author={Song, Rui and Liang, Chenwei and Cao, Hu and Yan, Zhiran and Zimmer, Walter and Gross, Markus and Festag, Andreas and Knoll, Alois},
  booktitle=CVPR,
  pages={17996--18006},
  year={2024}
}

@inproceedings{hu2023collaboration,
  title={Collaboration helps camera overtake lidar in 3d detection},
  author={Hu, Yue and Lu, Yifan and Xu, Runsheng and Xie, Weidi and Chen, Siheng and Wang, Yanfeng},
  booktitle=CVPR,
  pages={9243--9252},
  year={2023}
}

@inproceedings{arumugam2010davinci,
  title={DAvinCi: A cloud computing framework for service robots},
  author={Arumugam, Rajesh and Enti, Vikas Reddy and Bingbing, Liu and Xiaojun, Wu and Baskaran, Krishnamoorthy and Kong, Foong Foo and Kumar, A Senthil and Meng, Kang Dee and Kit, Goh Wai},
  booktitle=ICRA,
  pages={3084--3089},
  year={2010},
  organization={IEEE}
}

@article{riazuelo2014c2tam,
  title={C2tam: A cloud framework for cooperative tracking and mapping},
  author={Riazuelo, Luis and Civera, Javier and Montiel, JM Mart{\i}nez},
  journal={Robot. Auton. Syst.},
  volume={62},
  number={4},
  pages={401--413},
  year={2014},
  publisher={Elsevier}
}

@InProceedings{pmlr-v205-li23e,
  title = 	 {Multi-Robot Scene Completion: Towards Task-Agnostic Collaborative Perception},
  author =       {Li, Yiming and Zhang, Juexiao and Ma, Dekun and Wang, Yue and Feng, Chen},
  booktitle = 	 {Conf. Robot Learn},
  pages = 	 {2062--2072},
  year = 	 {2023},
  editor = 	 {Liu, Karen and Kulic, Dana and Ichnowski, Jeff},
  volume = 	 {205},
  series = 	 {Proceedings of Machine Learning Research},
  month = 	 {14--18 Dec},
  publisher =    {PMLR},
  url = 	 {https://proceedings.mlr.press/v205/li23e.html}
}

@misc{li2022learningdistilledcollaborationgraph,
      title={Learning Distilled Collaboration Graph for Multi-Agent Perception}, 
      author={Yiming Li and Shunli Ren and Pengxiang Wu and Siheng Chen and Chen Feng and Wenjun Zhang},
      year={2022},
      eprint={2111.00643},
      archivePrefix={arXiv},
note={arXiv:2111.00643},
      primaryClass={cs.CV},
      url={https://arxiv.org/abs/2111.00643}, 
}

@inproceedings{xu2022v2x,
  title={V2x-vit: Vehicle-to-everything cooperative perception with vision transformer},
  author={Xu, Runsheng and Xiang, Hao and Tu, Zhengzhong and Xia, Xin and Yang, Ming-Hsuan and Ma, Jiaqi},
  booktitle=ECCV,
  pages={107--124},
  year={2022},
  organization={Springer}
}

@misc{liu2023towards,
      title={Towards Vehicle-to-everything Autonomous Driving: A Survey on Collaborative Perception}, 
      author={Si Liu and Chen Gao and Yuan Chen and Xingyu Peng and Xianghao Kong and Kun Wang and Runsheng Xu and Wentao Jiang and Hao Xiang and Jiaqi Ma and Miao Wang},
      year={2023},
      eprint={2308.16714},
      archivePrefix={arXiv},
      primaryClass={cs.CV},
      url={https://arxiv.org/abs/2308.16714},
    note={arXiv:2308.16714}
}

@inproceedings{yin2021center,
  title={Center-based 3d object detection and tracking},
  author={Yin, Tianwei and Zhou, Xingyi and Krahenbuhl, Philipp},
  booktitle=CVPR,
  pages={11784--11793},
  year={2021}
}

@inproceedings{song2018design,
  title={Design and implementation of a pose estimation system based on visual fiducial features and multiple cameras},
  author={Song, Kai-Tai and Chang, Yueh Chuan},
  booktitle={Int. Autom. Control Conf.},
  pages={1--6},
  year={2018},
  organization={IEEE}
}

@article{li2023delving,
  title={Delving into the devils of bird’s-eye-view perception: A review, evaluation and recipe},
  author={Li, Hongyang and Sima, Chonghao and Dai, Jifeng and Wang, Wenhai and Lu, Lewei and Wang, Huijie and Zeng, Jia and Li, Zhiqi and Yang, Jiazhi and Deng, Hanming and others},
  journal=TPAMI,
  volume={46},
  number={4},
  pages={2151--2170},
  year={2023},
  publisher={IEEE}
}

@article{li2024bevformer,
  title={Bevformer: learning bird's-eye-view representation from lidar-camera via spatiotemporal transformers},
  author={Li, Zhiqi and Wang, Wenhai and Li, Hongyang and Xie, Enze and Sima, Chonghao and Lu, Tong and Yu, Qiao and Dai, Jifeng},
  journal=TPAMI,
  year={2024},
  publisher={IEEE}
}

@article{liang2022bevfusion,
  title={Bevfusion: A simple and robust lidar-camera fusion framework},
  author={Liang, Tingting and Xie, Hongwei and Yu, Kaicheng and Xia, Zhongyu and Lin, Zhiwei and Wang, Yongtao and Tang, Tao and Wang, Bing and Tang, Zhi},
  journal=NIPS,
  volume={35},
  pages={10421--10434},
  year={2022}
}

@inproceedings{philion2020lift,
  title={Lift, splat, shoot: Encoding images from arbitrary camera rigs by implicitly unprojecting to 3d},
  author={Philion, Jonah and Fidler, Sanja},
  booktitle=ECCV,
  pages={194--210},
  year={2020},
  organization={Springer}
}

@inproceedings{lang2019pointpillars,
  title={Pointpillars: Fast encoders for object detection from point clouds},
  author={Lang, Alex H and Vora, Sourabh and Caesar, Holger and Zhou, Lubing and Yang, Jiong and Beijbom, Oscar},
  booktitle=CVPR,
  pages={12697--12705},
  year={2019}
}

@inproceedings{hou2020multiview,
  title={Multiview detection with feature perspective transformation},
  author={Hou, Yunzhong and Zheng, Liang and Gould, Stephen},
  booktitle=ECCV,
  pages={1--18},
  year={2020},
  organization={Springer}
}

@article{fleuret2008multicamera,
  title={Multicamera people tracking with a probabilistic occupancy map},
  author={Fleuret, Francois and Berclaz, Jerome and Lengagne, Richard and Fua, Pascal},
  journal=TPAMI,
  volume={30},
  number={2},
  pages={267--282},
  year={2008},
  publisher={IEEE}
}

@inproceedings{roig2011conditional,
  title={Conditional random fields for multi-camera object detection},
  author={Roig, Gemma and Boix, Xavier and Shitrit, Horesh Ben and Fua, Pascal},
  booktitle=ICCV,
  pages={563--570},
  year={2011},
  organization={IEEE}
}

@inproceedings{zhou2020tracking,
  title={Tracking objects as points},
  author={Zhou, Xingyi and Koltun, Vladlen and Kr{\"a}henb{\"u}hl, Philipp},
  booktitle=ECCV,
  pages={474--490},
  year={2020},
  organization={Springer}
}

@inproceedings{chiu2024probabilistic,
  title={Probabilistic 3d multi-object cooperative tracking for autonomous driving via differentiable multi-sensor kalman filter},
  author={Chiu, Hsu-Kuang and Wang, Chien-Yi and Chen, Min-Hung and Smith, Stephen F},
  booktitle=ICRA,
  pages={18458--18464},
  year={2024},
  organization={IEEE}
}

@article{williams2015marginal,
  title={Marginal multi-Bernoulli filters: RFS derivation of MHT, JIPDA, and association-based MeMBer},
  author={Williams, Jason L},
  journal={IEEE Trans. Aerosp. Electron. Syst.},
  volume={51},
  number={3},
  pages={1664--1687},
  year={2015},
  publisher={IEEE}
}

@article{hu2024integrated,
  title={Integrated Detection and Tracking Framework for 3D Multi-Object Tracking in Vehicle-Infrastructure Cooperation.},
  author={Hu, Tao and Wang, Ping and Wang, Xinhong},
  journal={Int. J. Adv. Comput. Sci. Appl.},
  volume={15},
  number={11},
  year={2024}
}

@article{garcia2018poisson,
  title={Poisson multi-Bernoulli mixture filter: Direct derivation and implementation},
  author={Garc{\'\i}a-Fern{\'a}ndez, {\'A}ngel F and Williams, Jason L and Granstr{\"o}m, Karl and Svensson, Lennart},
  journal={IEEE Trans. Aerosp. Electron. Syst.},
  volume={54},
  number={4},
  pages={1883--1901},
  year={2018},
  publisher={IEEE}
}

@inproceedings{kim2021eagermot,
  title={Eagermot: 3d multi-object tracking via sensor fusion},
  author={Kim, Aleksandr and O{\v{s}}ep, Aljo{\v{s}}a and Leal-Taix{\'e}, Laura},
  booktitle=ICRA,
  pages={11315--11321},
  year={2021},
  organization={IEEE}
}

@inproceedings{chiu2021probabilistic,
  title={Probabilistic 3D multi-modal, multi-object tracking for autonomous driving},
  author={Chiu, Hsu-kuang and Li, Jie and Ambru{\c{s}}, Rare{\c{s}} and Bohg, Jeannette},
  booktitle=ICRA,
  pages={14227--14233},
  year={2021},
  organization={IEEE}
}

@inproceedings{weng20203d,
  title={3d multi-object tracking: A baseline and new evaluation metrics},
  author={Weng, Xinshuo and Wang, Jianren and Held, David and Kitani, Kris},
  booktitle=IROS,
  pages={10359--10366},
  year={2020},
  organization={IEEE}
}

@misc{chiu2020probabilistic3dmultiobjecttracking,
      title={Probabilistic 3D Multi-Object Tracking for Autonomous Driving}, 
      author={Hsu-kuang Chiu and Antonio Prioletti and Jie Li and Jeannette Bohg},
      year={2020},
      eprint={2001.05673},
      archivePrefix={arXiv},
      primaryClass={cs.CV},
      url={https://arxiv.org/abs/2001.05673},
note={arXiv:2001.05673}
}

@inproceedings{zhang2023motiontrack,
  title={Motiontrack: End-to-end transformer-based multi-object tracking with lidar-camera fusion},
  author={Zhang, Ce and Zhang, Chengjie and Guo, Yiluan and Chen, Lingji and Happold, Michael},
  booktitle=CVPR,
  pages={151--160},
  year={2023}
}

@article{zhao2024autonomous,
  title={Autonomous driving system: A comprehensive survey},
  author={Zhao, Jingyuan and Zhao, Wenyi and Deng, Bo and Wang, Zhenghong and Zhang, Feng and Zheng, Wenxiang and Cao, Wanke and Nan, Jinrui and Lian, Yubo and Burke, Andrew F},
  journal=ESA,
  volume={242},
  pages={122836},
  year={2024},
  publisher={Elsevier}
}

@article{badue2021self,
  title={Self-driving cars: A survey},
  author={Badue, Claudine and Guidolini, R{\^a}nik and Carneiro, Raphael Vivacqua and Azevedo, Pedro and Cardoso, Vinicius B and Forechi, Avelino and Jesus, Luan and Berriel, Rodrigo and Paixao, Thiago M and Mutz, Filipe and others},
  journal=ESA,
  volume={165},
  pages={113816},
  year={2021},
  publisher={Elsevier}
}

@article{paden2016survey,
  title={A survey of motion planning and control techniques for self-driving urban vehicles},
  author={Paden, Brian and {\v{C}}{\'a}p, Michal and Yong, Sze Zheng and Yershov, Dmitry and Frazzoli, Emilio},
  journal=TEV,
  volume={1},
  number={1},
  pages={33--55},
  year={2016},
  publisher={IEEE}
}

@article{teng2023motion,
  title={Motion planning for autonomous driving: The state of the art and future perspectives},
  author={Teng, Siyu and Hu, Xuemin and Deng, Peng and Li, Bai and Li, Yuchen and Ai, Yunfeng and Yang, Dongsheng and Li, Lingxi and Xuanyuan, Zhe and Zhu, Fenghua and others},
  journal=TEV,
  volume={8},
  number={6},
  pages={3692--3711},
  year={2023},
  publisher={IEEE}
}

@inproceedings{roselli2021solving,
  title={Solving the conflict-free electric vehicle routing problem using SMT solvers},
  author={Roselli, Sabino Francesco and Fabian, Martin and {\AA}kesson, Knut},
  booktitle={Mediterr. Conf. Control Autom.},
  pages={542--547},
  year={2021},
  organization={IEEE}
}

@article{roselli2022compositional,
  title={A compositional algorithm for the conflict-free electric vehicle routing problem},
  author={Roselli, Sabino Francesco and G{\"o}tvall, Per-Lage and Fabian, Martin and {\AA}kesson, Knut},
  journal={IEEE Trans. Autom. Sci. Eng.},
  volume={19},
  number={3},
  pages={1405--1421},
  year={2022},
  publisher={IEEE}
}

@article{simoens2018internet,
  title={The Internet of Robotic Things: A review of concept, added value and applications},
  author={Simoens, Pieter and Dragone, Mauro and Saffiotti, Alessandro},
  journal={Int. J. Adv. Robot. Syst.},
  volume={15},
  number={1},
  pages={1729881418759424},
  year={2018},
  publisher={Sage Publications Sage UK: London, England}
}

@unpublished{ahmed2025realistisk,
    author = {Ahmed, Hussein and Bodin, Linnea and Jansson, Emil and Nilsson, Lucas and Sj{\"o}stedt, Melker and Hafstad Wallander, David},
    title = {Realistisk tr{\"a}ningsdata med 3D Gaussian Splatting: Bygga digitala tvillingar f{\"o}r generering av ground-truth-data i Unreal Engine 5},
    note = 2025
}

@article{hu2012cloud,
  title={Cloud robotics: architecture, challenges and applications},
  author={Hu, Guoqiang and Tay, Wee Peng and Wen, Yonggang},
  journal={IEEE network},
  volume={26},
  number={3},
  pages={21--28},
  year={2012},
  publisher={IEEE}
}

@inproceedings{popolizio2024online,
  title={Online Conflict-Free Scheduling of Fleets of Autonomous Mobile Robots},
  author={Popolizio, Francesco and Vinetti, Martina and Combrink, Alvin and Roselli, Sabino Francesco and Fanti, Maria Pia and Fabian, Martin},
  booktitle=CASE,
  pages={3063--3068},
  year={2024},
  organization={IEEE}
}

@article{jun2022scheduling,
  title={Scheduling of autonomous mobile robots with conflict-free routes utilising contextual-bandit-based local search},
  author={Jun, Sungbum and Choi, Chul Hun and Lee, Seokcheon},
  journal={Int. J. Prod. Res.},
  volume={60},
  number={13},
  pages={4090--4116},
  year={2022},
  publisher={Taylor \& Francis}
}

@article{lin2022review,
  title={A review of path-planning approaches for multiple mobile robots},
  author={Lin, Shiwei and Liu, Ang and Wang, Jianguo and Kong, Xiaoying},
  journal={Machines},
  volume={10},
  number={9},
  pages={773},
  year={2022},
  publisher={MDPI}
}

@article{konstantakopoulos2022vehicle,
  title={Vehicle routing problem and related algorithms for logistics distribution: A literature review and classification},
  author={Konstantakopoulos, Grigorios D and Gayialis, Sotiris P and Kechagias, Evripidis P},
  journal={Oper. Res.},
  volume={22},
  number={3},
  pages={2033--2062},
  year={2022},
  publisher={Springer}
}

@Article{amr_pathplanning,
AUTHOR = {Karur, Karthik and Sharma, Nitin and Dharmatti, Chinmay and Siegel, Joshua E.},
TITLE = {A Survey of Path Planning Algorithms for Mobile Robots},
JOURNAL = {Vehicles},
VOLUME = {3},
YEAR = {2021},
NUMBER = {3},
PAGES = {448--468},
URL = {https://www.mdpi.com/2624-8921/3/3/27},
ISSN = {2624-8921},
DOI = {10.3390/vehicles3030027}
}

@INPROCEEDINGS{mpc-rl,
  author={Ceder, Kristian and Zhang, Ze and Burman, Adam and Kuangaliyev, Ilya and Mattsson, Krister and Nyman, Gabriel and Petersén, Arvid and Wisell, Lukas and Åkesson, Knut},
  booktitle=IROS, 
  title={Bird’s-Eye-View Trajectory Planning of Multiple Robots using Continuous Deep Reinforcement Learning and Model Predictive Control}, 
  year={2024},
  volume={},
  number={},
  pages={8002-8008},
  doi={10.1109/IROS58592.2024.10801434}}

@INPROCEEDINGS{mpc-dqn,
  author={Zhang, Ze and Cai, Yao and Ceder, Kristian and Enliden, Arvid and Eriksson, Ossian and Kylander, Soleil and Sridhara, Rajath and Åkesson, Knut},
  booktitle=CASE, 
  title={Collision-Free Trajectory Planning of Mobile Robots by Integrating Deep Reinforcement Learning and Model Predictive Control}, 
  year={2023},
  volume={},
  number={},
  pages={1-7},
  doi={10.1109/CASE56687.2023.10260515}}

@INPROCEEDINGS{dwa-rl,
  author={Patel, Utsav and Kumar, Nithish K Sanjeev and Sathyamoorthy, Adarsh Jagan and Manocha, Dinesh},
  booktitle=ICRA, 
  title={DWA-RL: Dynamically Feasible Deep Reinforcement Learning Policy for Robot Navigation among Mobile Obstacles}, 
  year={2021},
  volume={},
  number={},
  pages={6057-6063},
  doi={10.1109/ICRA48506.2021.9561462}}

@INPROCEEDINGS{d_star_improved,
  author={Guo, Jianming and Liu, Liang and Liu, Qing and Qu, Yongyu},
  booktitle={Int. Conf. Intell. Comput. Technol. Autom.}, 
  title={An Improvement of D* Algorithm for Mobile Robot Path Planning in Partial Unknown Environment}, 
  year={2009},
  volume={3},
  number={},
  pages={394-397},
  doi={10.1109/ICICTA.2009.561}}

@Article{teb_improved,
AUTHOR = {Wu, Jiafeng and Ma, Xianghua and Peng, Tongrui and Wang, Haojie},
TITLE = {An Improved Timed Elastic Band (TEB) Algorithm of Autonomous Ground Vehicle (AGV) in Complex Environment},
JOURNAL = {Sensors},
VOLUME = {21},
YEAR = {2021},
NUMBER = {24},
ARTICLE-NUMBER = {8312},
URL = {https://www.mdpi.com/1424-8220/21/24/8312},
PubMedID = {34960406},
ISSN = {1424-8220},
DOI = {10.3390/s21248312}
}

@inproceedings{xu2016multi,
  title={Multi-view people tracking via hierarchical trajectory composition},
  author={Xu, Yuanlu and Liu, Xiaobai and Liu, Yang and Zhu, Song-Chun},
  booktitle=CVPR,
  pages={4256--4265},
  year={2016}
}

@inproceedings{cheng2023rest,
  title={Rest: A reconfigurable spatial-temporal graph model for multi-camera multi-object tracking},
  author={Cheng, Cheng-Che and Qiu, Min-Xuan and Chiang, Chen-Kuo and Lai, Shang-Hong},
  booktitle=ICCV,
  pages={10051--10060},
  year={2023}
}

@article{ong2020bayesian,
  title={A Bayesian filter for multi-view 3D multi-object tracking with occlusion handling},
  author={Ong, Jonah and Vo, Ba-Tuong and Vo, Ba-Ngu and Kim, Du Yong and Nordholm, Sven},
  journal=TPAMI,
  volume={44},
  number={5},
  pages={2246--2263},
  year={2020},
  publisher={IEEE}
}

@inproceedings{teepe2024earlybird,
  title={EarlyBird: early-fusion for multi-view tracking in the bird's eye View},
  author={Teepe, Torben and Wolters, Philipp and Gilg, Johannes and Herzog, Fabian and Rigoll, Gerhard},
  booktitle=WACV,
  pages={102--111},
  year={2024}
}

@inproceedings{meinhardt2022trackformer,
  title={Trackformer: Multi-object tracking with transformers},
  author={Meinhardt, Tim and Kirillov, Alexander and Leal-Taixe, Laura and Feichtenhofer, Christoph},
  booktitle=CVPR,
  pages={8844--8854},
  year={2022}
}

@ARTICLE{2023_dong,
  author={Dong, Lu and He, Zichen and Song, Chunwei and Sun, Changyin},
  journal={J. Syst. Eng. Electron.}, 
  title={A review of mobile robot motion planning methods: from classical motion planning workflows to reinforcement learning-based architectures}, 
  year={2023},
  volume={34},
  number={2},
  pages={439-459},
  doi={10.23919/JSEE.2023.000051}}

@INPROCEEDINGS{robusttemporalbelta2015,
  author={Sadraddini, Sadra and Belta, Calin},
  booktitle={Annu. Allerton Conf. Commun. Control Comput.}, 
  title={Robust temporal logic model predictive control}, 
  year={2015},
  volume={},
  number={},
  pages={772-779},
  doi={10.1109/ALLERTON.2015.7447084}}

@ARTICLE{lindemanncbf2019,
  author={Lindemann, Lars and Dimarogonas, Dimos V.},
  journal={IEEE Control Syst. Lett.}, 
  title={Control Barrier Functions for Signal Temporal Logic Tasks}, 
  year={2019},
  volume={3},
  number={1},
  pages={96-101},
  doi={10.1109/LCSYS.2018.2853182}}

@ARTICLE{Charitidoudmpc2024,
  author={Charitidou, Maria and Dimarogonas, Dimos V.},
  journal={IEEE Control Syst. Lett.}, 
  title={Distributed MPC With Continuous-Time STL Constraint Satisfaction Guarantees}, 
  year={2024},
  volume={8},
  number={},
  pages={211-216},
  doi={10.1109/LCSYS.2024.3361971}}

@inproceedings{scenic2019,
author = {Fremont, Daniel J. and Dreossi, Tommaso and Ghosh, Shromona and Yue, Xiangyu and Sangiovanni-Vincentelli, Alberto L. and Seshia, Sanjit A.},
title = {Scenic: a language for scenario specification and scene generation},
year = {2019},
isbn = {9781450367127},
publisher = {Association for Computing Machinery},
booktitle = {ACM SIGPLAN Conf. Program. Lang. Des. Implement.},
pages = {63–78},
numpages = {16},
location = {Phoenix, AZ, USA},
series = {PLDI 2019}
}

@article{donze2010breach,
  title={Breach, A Toolbox for Verification and Parameter Synthesis of Hybrid Systems},
  author={Donz{\'e}, Alexandre},
  journal=CAV,
  year={2010},
  volume={6174},
  series={Lecture Notes in Computer Science},
  pages={167--170},
  publisher={Springer},
  doi={10.1007/978-3-642-14295-6_17}
}

@inproceedings{verifaicav19,
  author    = {Tommaso Dreossi and
               Daniel J. Fremont and
               Shromona Ghosh and
               Edward Kim and
               Hadi Ravanbakhsh and
               Marcell Vazquez{-}Chanlatte and
               Sanjit A. Seshia},
  title     = {{VerifAI:} {A} Toolkit for the Formal Design and Analysis of Artificial Intelligence-Based Systems},
  booktitle = CAV,
  month = jul,
  year = {2019}
}

@ARTICLE{trajectory2025review,
  author={Souza de Oliveira, Cristiano and Toledo, Rafael De S. and Tulux Oliveira Victorio, Vitor H. and von Wangenheim, Aldo},
  journal={IEEE Access}, 
  title={Trajectory Planning for Autonomous Cars in Low-Structured and Unstructured Environments: A Systematic Review}, 
  year={2025},
  volume={13},
  number={},
  pages={48841-48871},
  doi={10.1109/ACCESS.2025.3551453}}

@Article{uav2025survey,
AUTHOR = {Zhou, Yuquan and Yan, Li and Han, Yaxi and Xie, Hong and Zhao, Yinghao},
TITLE = {A Survey on the Key Technologies of UAV Motion Planning},
JOURNAL = {Drones},
VOLUME = {9},
YEAR = {2025},
NUMBER = {3},
ARTICLE-NUMBER = {194},
URL = {https://www.mdpi.com/2504-446X/9/3/194},
ISSN = {2504-446X},
DOI = {10.3390/drones9030194}
}

@article{safe2024survey,
author = {Hwang, Sunwoo and Jang, Inkyu and Kim, Dabin},
year = {2024},
month = {10},
pages = {2955-2969},
title = {Safe Motion Planning and Control for Mobile Robots: A Survey},
volume = {22},
journal = {Int. J. Control Autom. Syst.},
doi = {10.1007/s12555-024-0784-5}
}

@ARTICLE{trajectory2024survey,
  author={Guo, Yuqing and Guo, Zelin and Wang, Yazhou and Yao, Danya and Li, Bai and Li, Li},
  journal=TEV, 
  title={A Survey of Trajectory Planning Methods for Autonomous Driving—Part I: Unstructured Scenarios}, 
  year={2024},
  volume={9},
  number={9},
  pages={5407-5434},
  doi={10.1109/TIV.2023.3337318}}

@article{intelligent2022review,
  title={Motion planning and control for mobile robot navigation using machine learning: a survey},
  author={Xiao, Xuesu and Liu, Bo and Warnell, Garrett and Stone, Peter},
  journal={Autonomous Robots},
  volume={46},
  number={5},
  pages={569--597},
  year={2022},
  publisher={Springer}
}

@ARTICLE{888718,
  author={Zhang, Z.},
  journal=TPAMI, 
  title={A flexible new technique for camera calibration}, 
  year={2000},
  volume={22},
  number={11},
  pages={1330-1334},
  doi={10.1109/34.888718}}

@ARTICLE{cvm_2020,
  title={What the Constant Velocity Model Can Teach Us About Pedestrian Motion Prediction}, 
  author={Schöller, Christoph and Aravantinos, Vincent and Lay, Florian and Knoll, Alois},
  journal={IEEE Robot. Autom. Lett.}, 
  year={2020},
  volume={5},
  number={2},
  pages={1696-1703},
  doi={10.1109/LRA.2020.2969925}
}

@article{sf_1995,
  title={Social force model for pedestrian dynamics},
  author={Helbing, Dirk and Molnar, Peter},
  journal={Physical review E},
  volume={51},
  number={5},
  pages={4282},
  year={1995},
  publisher={APS}
}

@INPROCEEDINGS{rvo_2008,
  author={van den Berg, Jur and Ming Lin and Manocha, Dinesh},
  booktitle=ICRA, 
  title={Reciprocal Velocity Obstacles for real-time multi-agent navigation}, 
  year={2008},
  volume={},
  number={},
  pages={1928-1935},
  doi={10.1109/ROBOT.2008.4543489}
}

@inproceedings{socialgan_2018_Gupta,
  title={Social {GAN}: Socially acceptable trajectories with generative adversarial networks},
  author={Gupta, Agrim and Johnson, Justin and Fei-Fei, Li and Savarese, Silvio and Alahi, Alexandre},
  booktitle=CVPR,
  pages={2255--2264},
  year={2018}
}

@inproceedings{fang_2020_tpnet,
  title={{TPNet}: Trajectory proposal network for motion prediction},
  author={Fang, Liangji and Jiang, Qinhong and Shi, Jianping and Zhou, Bolei},
  booktitle=CVPR,
  pages={6797--6806},
  year={2020}
}

@inproceedings{overcoming_2019_Makansi,
    title = {Overcoming Limitations of Mixture Density Networks: A Sampling and Fitting Framework for Multimodal Future Prediction},
    author = {Makansi, Osama and Ilg, Eddy and Cicek, Ozgun and Brox, Thomas},
    booktitle = CVPR,
    year = {2019},
}

@INPROCEEDINGS{ynet_2021_Mangalam,
  title={From Goals, Waypoints \& Paths To Long Term Human Trajectory Forecasting}, 
  author={Mangalam, Karttikeya and An, Yang and Girase, Harshayu and Malik, Jitendra},
  booktitle=ICCV, 
  year={2021},
  pages={15213-15222},
  doi={10.1109/ICCV48922.2021.01495}
}

@ARTICLE{ze_2025,
  title={Future-Oriented Navigation: Dynamic Obstacle Avoidance With One-Shot Energy-Based Multimodal Motion Prediction}, 
  author={Zhang, Ze and Hess, Georg and Hu, Junjie and Dean, Emmanuel and Svensson, Lennart and Åkesson, Knut},
  journal={IEEE Robot. Autom. Lett.}, 
  year={2025},
  volume={10},
  number={8},
  pages={8043-8050},
  doi={10.1109/LRA.2025.3575969}
}

@article{rittel1973dilemmas,
  title={Dilemmas in a general theory of planning},
  author={Rittel, Horst WJ and Webber, Melvin M},
  journal={Policy sciences},
  volume={4},
  number={2},
  pages={155--169},
  year={1973},
  publisher={Springer}
}

@software{RerunSDK,
  title = {Rerun: A Visualization SDK for Multimodal Data},
  author = {{Rerun Development Team}},
  url = {https://www.rerun.io},
  version = {insert version number},
  date = {insert date of usage},
  year = {2024},
  publisher = {{Rerun Technologies AB}},
  address = {Online},
  note = {Available from https://www.rerun.io/ and https://github.com/rerun-io/rerun}
}

@article{macenski2023survey,
      title={From the desks of ROS maintainers: A survey of modern \& capable mobile robotics algorithms in the robot operating system 2},
      author={Macenski, Steve and Moore, Tom and Lu, David V. and Merzlyakov, Alexey and Ferguson, Michael},
      year={2023},
      journal = {Robot. Auton. Syst.}
}

@article{macenski2021slam,
  title={SLAM Toolbox: SLAM for the dynamic world},
  author={Macenski, Steve and Jambrecic, Ivona},
  journal={J. Open Source Softw.},
  volume={6},
  number={61},
  pages={2783},
  year={2021}
}

@article{saha2018comprehensive,
  title={A comprehensive survey of recent trends in cloud robotics architectures and applications},
  author={Saha, Olimpiya and Dasgupta, Prithviraj},
  journal={Robotics},
  volume={7},
  number={3},
  pages={47},
  year={2018},
  publisher={MDPI}
}

@article{cadena2017past,
  title={Past, present, and future of simultaneous localization and mapping: Toward the robust-perception age},
  author={Cadena, Cesar and Carlone, Luca and Carrillo, Henry and Latif, Yasir and Scaramuzza, Davide and Neira, Jos{\'e} and Reid, Ian and Leonard, John J},
  journal={IEEE Trans. Robot.},
  volume={32},
  number={6},
  pages={1309--1332},
  year={2017},
  publisher={IEEE}
}

@article{grisetti2007improved,
  title={Improved techniques for grid mapping with rao-blackwellized particle filters},
  author={Grisetti, Giorgio and Stachniss, Cyrill and Burgard, Wolfram},
  journal={IEEE Trans. Robot.},
  volume={23},
  number={1},
  pages={34--46},
  year={2007},
  publisher={IEEE}
}

@inproceedings{konolige2010efficient,
  title={Efficient sparse pose adjustment for 2D mapping},
  author={Konolige, Kurt and Grisetti, Giorgio and K{\"u}mmerle, Rainer and Burgard, Wolfram and Limketkai, Benson and Vincent, Regis},
  booktitle=IROS,
  pages={22--29},
  year={2010},
  organization={IEEE}
}

@inproceedings{hess2016real,
  title={Real-time loop closure in 2D LIDAR SLAM},
  author={Hess, Wolfgang and Kohler, Damon and Rapp, Holger and Andor, Daniel},
  booktitle=ICRA,
  pages={1271--1278},
  year={2016},
  organization={IEEE}
}

@article{zou2021comparative,
  title={A comparative analysis of LiDAR SLAM-based indoor navigation for autonomous vehicles},
  author={Zou, Qin and Sun, Qin and Chen, Long and Nie, Bu and Li, Qingquan},
  journal={IEEE Trans. Intell. Transp. Syst.},
  volume={23},
  number={7},
  pages={6907--6921},
  year={2021},
  publisher={IEEE}
}

@article{macario2022comprehensive,
  title={A comprehensive survey of visual slam algorithms},
  author={Macario Barros, Andr{\'e}a and Michel, Maugan and Moline, Yoann and Corre, Gwenol{\'e} and Carrel, Fr{\'e}d{\'e}rick},
  journal={Robotics},
  volume={11},
  number={1},
  pages={24},
  year={2022},
  publisher={MDPI}
}

@article{karrer2018cvi,
  title={CVI-SLAM—collaborative visual-inertial SLAM},
  author={Karrer, Marco and Schmuck, Patrik and Chli, Margarita},
  journal={IEEE Robot. Autom. Lett.},
  volume={3},
  number={4},
  pages={2762--2769},
  year={2018},
  publisher={IEEE}
}

@article{campos2021orb,
  title={Orb-slam3: An accurate open-source library for visual, visual--inertial, and multimap slam},
  author={Campos, Carlos and Elvira, Richard and Rodr{\'\i}guez, Juan J G{\'o}mez and Montiel, Jos{\'e} MM and Tard{\'o}s, Juan D},
  journal={IEEE Trans. Robot.},
  volume={37},
  number={6},
  pages={1874--1890},
  year={2021},
  publisher={IEEE}
}

@ARTICLE{selfattention_2024_yang,
  author={Yang, Changzhi and Pan, Huihui and Sun, Weichao and Gao, Huijun},
  journal={IEEE Trans. Artif. Intell.}, 
  title={Social Self-Attention Generative Adversarial Networks for Human Trajectory Prediction}, 
  year={2024},
  volume={5},
  number={4},
  pages={1805-1815},
  doi={10.1109/TAI.2023.3299899}
}

@inproceedings{Trajectronpp_2020_Salzmann,
  title={Trajectron++: Dynamically-feasible trajectory forecasting with heterogeneous data},
  author={Salzmann, Tim and Ivanovic, Boris and Chakravarty, Punarjay and Pavone, Marco},
  booktitle=ECCV,
  pages={683--700},
  year={2020},
  organization={Springer}
}

@INPROCEEDINGS {end2end_2022_guo,
    title = {End-to-End Trajectory Distribution Prediction Based on Occupancy Grid Maps},
    author = {K. Guo and W. Liu and J. Pan},
    booktitle = CVPR,
    year = {2022},
    volume = {},
    issn = {},
    pages = {2232-2241},
}

@article{bogyrbayeva2024machine,
  title={Machine learning to solve vehicle routing problems: A survey},
  author={Bogyrbayeva, Aigerim and Meraliyev, Meraryslan and Mustakhov, Taukekhan and Dauletbayev, Bissenbay},
  journal={IEEE Trans. Intell. Transp. Syst.},
  volume={25},
  number={6},
  pages={4754--4772},
  year={2024},
  publisher={IEEE}
}

@incollection{alenljung2023towards,
  title={Towards a Framework of Human-Robot Interaction Strategies from an Operator 5.0 Perspective},
  author={Alenljung, Beatrice and Lindblom, Jessica and Zaragoza-Sundqvist, Maximiliano and Hanna, Atieh},
  booktitle={Adv. Manuf. Technol. XXXVI},
  pages={81--86},
  year={2023},
  publisher={IOS Press}
}

@article{lindblom2020anemone,
  title={The ANEMONE: theoretical foundations for UX evaluation of action and intention recognition in human-robot interaction},
  author={Lindblom, Jessica and Alenljung, Beatrice},
  journal={Sensors},
  volume={20},
  number={15},
  pages={4284},
  year={2020},
  publisher={MDPI}
}

@book{ghallab2004automated,
  title={Automated Planning: theory and practice},
  author={Ghallab, Malik and Nau, Dana and Traverso, Paolo},
  year={2004},
  publisher={Elsevier}
}

@article{mcdermott1998pddl,
  title={Pddl| the planning domain definition language},
  author={Aeronautiques, Constructions and Howe, Adele and Knoblock, Craig and McDermott, ISI Drew and Ram, Ashwin and Veloso, Manuela and Weld, Daniel and Sri, David Wilkins and Barrett, Anthony and Christianson, Dave and others},
  journal={Technical Report, Tech. Rep.},
  year={1998}
}

@article{fox2003pddl21,
  title={PDDL2. 1: An extension to PDDL for expressing temporal planning domains},
  author={Fox, Maria and Long, Derek},
  journal=JAIR,
  volume={20},
  pages={61--124},
  year={2003}
}

@article{paton1999active,
  title={Active database systems},
  author={Paton, Norman W and Diaz, Oscar},
  journal={ACM Comput. Surv.},
  volume={31},
  number={1},
  pages={63--103},
  year={1999},
  publisher={ACM New York, NY, USA}
}

@article{skoldstam2008supervisory,
  title={Supervisory control applied to automata extended with variables-revised},
  author={Sk{\"o}ldstam, Markus and {\AA}kesson, Knut and Fabian, Martin},
  journal={Relat{\'o}rio t{\'e}cnico, Goteborg: Chalmers University of Technology},
  year={2008}
}

@book{russell2016artificial,
  title={Artificial intelligence: a modern approach},
  author={Russell, Stuart J and Norvig, Peter},
  year={2016},
  edition   = {3},
  publisher={pearson}
}

@article{bryant1986graph,
  title={Graph-based algorithms for boolean function manipulation},
  author={Bryant, Randal E},
  journal={IEEE Trans. Comput.},
  volume={100},
  number={8},
  pages={677--691},
  year={1986},
  publisher={IEEE}
}

@inproceedings{kautz1992planning,
  title={Planning as Satisfiability.},
  author={Kautz, Henry A and Selman, Bart and others},
  booktitle={ECAI},
  volume={92},
  pages={359--363},
  year={1992},
  organization={Citeseer}
}

@inproceedings{pnueli1977temporal,
  title={The temporal logic of programs},
  author={Pnueli, Amir},
  booktitle={IEEE Annu. Symp. Found. Comput. Sci.},
  pages={46--57},
  year={1977},
  organization={ieee}
}

@book{baier2008principles,
  title={Principles of model checking},
  author={Baier, Christel and Katoen, Joost-Pieter},
  year={2008},
  publisher={MIT press}
}

@article{amir2008learning,
  title={Learning partially observable deterministic action models},
  author={Amir, Eyal and Chang, Allen},
  journal=JAIR,
  volume={33},
  pages={349--402},
  year={2008}
}

@ARTICLE{ashfaq2022learning,

  author={Farooqui, Ashfaq and Claase, Ramon Tijsse and Fabian, Martin},

  journal={IEEE Trans. Autom. Sci. Eng.}, 

  title={On Active Learning for Supervisor Synthesis}, 

  year={2024},

  volume={21},

  number={1},

  pages={78-90},

  doi={10.1109/TASE.2022.3216759}}

@INPROCEEDINGS{pathplanning2024review,
  author={Alarabi, Saleh and Santora, Michael},
  booktitle={Proc. 9th Asia-Pacific Conf. Intell. Robot Syst.}, 
  title={Review:Path Planning Techniques for Automated Guided Vehicles (AGVs)}, 
  year={2024},
  volume={},
  number={},
  pages={33-38},
  doi={10.1109/ACIRS62330.2024.10684912}}

@article{bai2023analytics,
  title={Analytics and machine learning in vehicle routing research},
  author={Bai, Ruibin and Chen, Xinan and Chen, Zhi-Long and Cui, Tianxiang and Gong, Shuhui and He, Wentao and Jiang, Xiaoping and Jin, Huan and Jin, Jiahuan and Kendall, Graham and others},
  journal={Int. J. Prod. Res.},
  volume={61},
  number={1},
  pages={4--30},
  year={2023},
  publisher={Taylor \& Francis}
}

@article{silva2019reinforcement,
  title={A reinforcement learning-based multi-agent framework applied for solving routing and scheduling problems},
  author={Silva, Maria Am{\'e}lia Lopes and de Souza, S{\'e}rgio Ricardo and Souza, Marcone Jamilson Freitas and Bazzan, Ana L{\'u}cia C},
  journal=ESA,
  volume={131},
  pages={148--171},
  year={2019},
  publisher={Elsevier}
}

@article{zhang2020multi,
  title={Multi-vehicle routing problems with soft time windows: A multi-agent reinforcement learning approach},
  author={Zhang, Ke and He, Fang and Zhang, Zhengchao and Lin, Xi and Li, Meng},
  journal={Transportation Research Part C: Emerging Technologies},
  volume={121},
  pages={102861},
  year={2020},
  publisher={Elsevier}
}

@misc{safexplain_d33_2025,
	title = {D3.3: {Final} proofs-of-concept, arguments, and {DL} components and libraries},
	publisher = {Deliverable of the HEU SAFEXPLAIN project, Grant Agreement No. 101069595},
	author = {{SAFEXPLAIN}},
	year = {2025},
    url = {https://safexplain.eu/deliverables},
  howpublished = {\url{https://safexplain.eu/deliverables}},

}

@misc{safexplain_d21_2024,
	title = {D2.1: {SAFEXPLAIN} {Safety} {Lifecycle} {Considerations}},
	publisher = {Deliverable of the HEU SAFEXPLAIN  project, Grant Agreement No. 101069595},
	author = {SAFEXPLAIN},
	year = {2024},
    url = {https://safexplain.eu/deliverables},
  howpublished = {\url{https://safexplain.eu/deliverables}},

}

@misc{sotif_2022,
	title = {Road vehicles — {Safety} of the intended functionality ({ISO} {Standard} {No}. 21448:2022)},
	shorttitle = {{SOTIF}},
	url = {https://www.iso.org/standard/77490.html},
	author = {{International Organization for Standardization}},
	year = {2022},
}

@techreport{tr5469,
	title = {Artificial intelligence — {Functional} safety and {AI} systems ({ISO} {Standard} {No}. 5469:2024)},
    institution = {ISO},
	url = {https://www.iso.org/standard/81283.html},
	author = {{International Organization for Standardization}},
	year = {2024},
}

@misc{hawkins_guidance_2021,
	title = {Guidance on the {Assurance} of {Machine} {Learning} in {Autonomous} {Systems} ({AMLAS})},
	url = {http://arxiv.org/abs/2102.01564},
	urldate = {2021-09-22},
	author = {Hawkins, Richard and Paterson, Colin and Picardi, Chiara and Jia, Yan and Calinescu, Radu and Habli, Ibrahim},
	month = feb,
	year = {2021},
	note = {arXiv:2102.01564},
}

@article{borg_ergo_2023,
	title = {Ergo, {SMIRK} is safe: a safety case for a machine learning component in a pedestrian automatic emergency brake system},
	issn = {1573-1367},
	url = {https://doi.org/10.1007/s11219-022-09613-1},
	doi = {10.1007/s11219-022-09613-1},
	journal = {Softw. Qual. J.},
	author = {Borg, Markus and Henriksson, Jens and Socha, Kasper and Lennartsson, Olof and Sonnsjö Lönegren, Elias and Bui, Thanh and Tomaszewski, Piotr and Sathyamoorthy, Sankar Raman and Brink, Sebastian and Helali Moghadam, Mahshid},
	month = mar,
	year = {2023},
}
}

\end{document}